\documentclass[fleqn,twoside]{article}
\usepackage{espcrc1}

\newcommand{\ttbs}{\char'134}
\newcommand{\AmS}{{\protect\the\textfont2
  A\kern-.1667em\lower.5ex\hbox{M}\kern-.125emS}}

\title{Towards Multistage Design of  Modular Systems}

\author{Mark Sh. Levin
\thanks{
 Mark Sh. Levin:~
  http://www.mslevin.iitp.ru;
 email: mslevin@acm.org
  }
  }

\begin{document}

\maketitle

\begin{abstract}
 The paper describes
 multistage design of composite (modular)
 systems (i.e., design of a system trajectory).
%
%
 This design process consists of the following:
 (i) definition of a set of time/logical points;
 (ii) modular design of the system for each time/logical point
 (e.g., on the basis of combinatorial synthesis as
 hierarchical morphological design
  or multiple choice problem)
 to obtain several system solutions;
 (iii) selection of the system solution for each time/logical point
 while taking into account
 their quality and the quality of compatibility between neighbor
 selected system solutions
 (here, combinatorial synthesis is used as well).
 Mainly,
 the examined time/logical points are based on a time chain.
 In addition, two complicated cases are considered:
 (a) the examined logical points are based on a tree-like structure,
 (b) the examined logical points are based on a digraph.
%
%
 Numerical examples illustrate the approach.

~~


\vspace{1pc}
\end{abstract}



\newcounter{cms}
\setlength{\unitlength}{1mm}

\section{Introduction}

 In recent decades,
 the significance of modular (multi-component) systems has been increased
 (e.g., \cite{bald00,dah01,garud09,hua98,jose05,lev98,lev06,lev12morph,lev13intro,ulrich94,voss09}).
 This paper addresses
 procedures for
 multistage design of hierarchical modular systems.
 As a result, a system trajectory can be obtained.
%
%
 It is assumed the following:
 the considered hierarchical modular system can be represented as
 a morphological structure:
  tree-like system structure,
 design alternatives (DAs) for leaf nodes of the structure,
 estimates of DAs and their compatibility
 (e.g., \cite{lev98,lev06,lev12morph,lev12hier}).
 In addition, it is necessary to consider
 a top-level structure/network
 (i.e., a set of elements as time/logical points/states and a structure over
 the elements), including the following  four basic types
 of the structure:
 (a) chain,
 (b) tree,
 (c) acyclic directed graph (digraph),
 and
 (d) general digraph.
 Thus, a two-level model
   is examined (Fig. 1):~

 (1) top-level network/graph ~\(G = (H,V)\), where
 \(H\) corresponds to a set of node (time/logical points or states),
 \(V\) corresponds to a set of arcs;

 (2) morphological structure for each  node \(\mu \in H\):~
 \(\Lambda^{\mu}\).

\begin{center}
\begin{picture}(90,32)
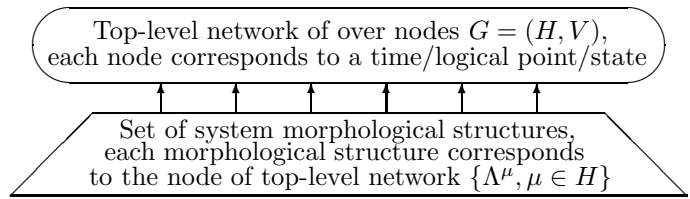
\put(00,00){\makebox(0,0)[bl]{Fig. 1. Two-level model for
 multistage system trajectories}}

\put(45,25){\oval(84,10)}

\put(11,25){\makebox(0,8)[bl]{Top-level network of over nodes
\(G=(H,V)\),}}

\put(05,21.5){\makebox(0,8)[bl]{each node corresponds to a
 time/logical point/state}}


\put(20,16){\vector(0,1){4}} \put(30,16){\vector(0,1){4}}
\put(40,16){\vector(0,1){4}}

\put(50,16){\vector(0,1){4}}

\put(60,16){\vector(0,1){4}} \put(70,16){\vector(0,1){4}}


\put(00,05){\line(1,0){90}} \put(11,16){\line(1,0){68}}
\put(00,05){\line(1,1){11}} \put(90,05){\line(-1,1){11}}

\put(14.5,12){\makebox(0,8)[bl]{Set of system morphological
 structures,}}

\put(12.5,9){\makebox(0,8)[bl]{each morphological structure
 corresponds}}

\put(10.5,5.5){\makebox(0,8)[bl]{to the node of top-level
 network \(\{ \Lambda^{\mu}, \mu \in H \}\)}}

\end{picture}
\end{center}

 Table 1 contains a brief description of the basic considered frameworks/problems.

%
 Generally,
 this kind of the design process consists of the following:
 (i) definition of a set of  time/logical points;
 (ii) definition of the structure over the points above
 (i.e., chain, tree, digraph);
 (iii) modular design of the system for each  point
 (e.g., on the basis of combinatorial synthesis as
 multiple choice problem  or hierarchical morphological design)
 to obtain several system solutions;
 (iv) selection of the system solution for each point
 while taking into account
 their quality and the quality of compatibility between neighbor
 selected system solutions
 (here, combinatorial synthesis is used as well).

  Mainly, a chain of time points is considered in
  the above-mentioned design scheme.
 In addition, two complicated cases are considered:
 (a) the examined logical/time points are based on a tree-like structure,
 (b) the examined logical/time points are based on a digraph.
%
%
 Numerical examples illustrate the approach.
 An illustration for
 using
  the approach to multiple domain problems is presented.

\begin{center}
\begin{picture}(115,138)
\put(08,134){\makebox(0,0)[bl]{Table 1. Some basic types of system
 trajectory design problems}}

\put(00,00){\line(1,0){115}} \put(00,118){\line(1,0){115}}
\put(00,132){\line(1,0){115}}

\put(00,00){\line(0,1){132}} \put(21,00){\line(0,1){132}}
\put(79,00){\line(0,1){132}} \put(115,00){\line(0,1){132}}

\put(01,128){\makebox(0,0)[bl]{Type of }}
\put(01,124){\makebox(0,0)[bl]{top-level}}
\put(01,120.5){\makebox(0,0)[bl]{network}}

\put(22,127.8){\makebox(0,0)[bl]{Type of design frameworks/}}
\put(22,124){\makebox(0,0)[bl]{problems}}

\put(80,127.5){\makebox(0,0)[bl]{Applied}}
\put(80,124.5){\makebox(0,0)[bl]{examples}}


\put(01,113){\makebox(0,0)[bl]{1.Chain }}

\put(22,113){\makebox(0,0)[bl]{(1.1) Series system
trajectory(ies)}}

\put(22,109){\makebox(0,0)[bl]{(1.2) Series-parallel system }}
\put(22,105){\makebox(0,0)[bl]{trajectory(ies)}}

\put(80,113){\makebox(0,0)[bl]{1.Series and parallel-}}
\put(80,109){\makebox(0,0)[bl]{series strategies for}}
\put(80,105){\makebox(0,0)[bl]{ranking/sorting}}
\put(80,101){\makebox(0,0)[bl]{\cite{lev98,lev12b} }}

\put(80,97){\makebox(0,0)[bl]{2.Series strategy for}}
\put(80,93){\makebox(0,0)[bl]{Web-based system}}
\put(80,89){\makebox(0,0)[bl]{(provider) \cite{lev11inf} }}

\put(80,85){\makebox(0,0)[bl]{3.Team design, }}
\put(80,81){\makebox(0,0)[bl]{this paper }}


\put(01,76.5){\makebox(0,0)[bl]{2.Tree-like}}
\put(04,72.5){\makebox(0,0)[bl]{structure}}

\put(22,76){\makebox(0,0)[bl]{(2.1) Tree-like system trajectory}}
\put(22,72){\makebox(0,0)[bl]{(2.2) Extension of the top-level
net-}}

\put(22,68){\makebox(0,0)[bl]{work by 'analysis' nodes; design}}
\put(22,64){\makebox(0,0)[bl]{of tree-like system
 trajectories}}

\put(80,76.5){\makebox(0,0)[bl]{Medical treatment,}}
\put(80,72){\makebox(0,0)[bl]{this paper}}


\put(01,59){\makebox(0,0)[bl]{3.Acyclic}}
\put(04,55.5){\makebox(0,0)[bl]{directed}}
\put(04,51){\makebox(0,0)[bl]{graph }}

\put(22,59){\makebox(0,0)[bl]{(3.1) Design of a route (e.g.,
chain, }}

\put(22,55){\makebox(0,0)[bl]{tree, parallel-series graph);}}
\put(22,51){\makebox(0,0)[bl]{design of corresponding system}}
\put(22,47){\makebox(0,0)[bl]{trajectory(ies)}}
\put(22,43){\makebox(0,0)[bl]{(3.2) Design of a spanning tree;}}
\put(22,39){\makebox(0,0)[bl]{solving the problem 2.2}}

\put(80,59.5){\makebox(0,0)[bl]{Example,}}
\put(80,55){\makebox(0,0)[bl]{this paper}}


\put(01,34.5){\makebox(0,0)[bl]{4.General}}
\put(04,30.5){\makebox(0,0)[bl]{directed }}
\put(04,26){\makebox(0,0)[bl]{graph }}

\put(22,34){\makebox(0,0)[bl]{(4.1) Design of a route (e.g.,
chain, }}

\put(22,30){\makebox(0,0)[bl]{tree, parallel-series graph);}}
\put(22,26){\makebox(0,0)[bl]{design of corresponding system}}
\put(22,22){\makebox(0,0)[bl]{trajectory(ies)}}

\put(22,18){\makebox(0,0)[bl]{(4.2) Design of a spanning tree;}}
\put(22,14){\makebox(0,0)[bl]{solving the problem 2.2}}

\put(22,10){\makebox(0,0)[bl]{(4.3) Design of a more simple}}
\put(22,06){\makebox(0,0)[bl]{(by structure) spanning network; }}
\put(22,02){\makebox(0,0)[bl]{solving problems 4.1 or 4.2}}


\put(80,34.5){\makebox(0,0)[bl]{Simplified illustration,}}
\put(80,30){\makebox(0,0)[bl]{this paper}}

\end{picture}
\end{center}

\section{Brief Description of Combinatorial Synthesis}

 Generally, combinatorial synthesis of modular systems can be
 based on multiple choice problem or
 hierarchical multicriteria morphological design (HMMD) approach
 (e.g., \cite{lev98,lev06,lev12morph,lev12a}.
 Here, HMMD is used.
%
%
 In HMMD approach,
  the resultant solution is composed from design alternatives
 (DAs) for system parts/components
 while taking into account quality if their interconnection (IC).
%
%
 In the basic version of HMMD, the following ordinal scales are used:
 (1) ordinal scale for quality of system components (or priority)
 (\(\iota = \overline{1,l}\);
      \(1\) corresponds to the best one);
 (2) scale for system quality while taking into account
 system components ordinal estimates
 and ordinal compatibility estimates between the system components
  (\(w=\overline{0,\nu}\); \(\nu\) corresponds to the best level).

 For the system consisting of \(m\) parts/components,
  a discrete space (poset, lattice) of the system quality (excellence) on the basis of the
 following vector is used:
 ~\(N(S)=(w(S);n(S))\),
 ~where \(w(S)\) is the minimum of pairwise compatibility
 between DAs which correspond to different system components,
 ~\(n(S)=(\eta_{1},...,\eta_{r},...,\eta_{k})\),
 ~where ~\(\eta_{r}\) is the number of DAs of the \(r\)th quality in ~\(S\)
 ~(\(\sum^{k}_{r=1} n_{r} = m \)).
 The optimization problem is:~~
 \(max ~N(S), ~~max ~w(S), ~~w(S) \geq 0\).

\section{General Schemes}

 The solving two-level framework  for
  multistage design or design of system trajectory is the
  following (Fig. 2)
 (e.g., \cite{lev98,lev06}):

~~

 {\it Phase 0.} Generation of general structure of the design
 problem:
  ~{\it 0.1.} generation of time/logical points,
  ~{\it 0.2.} generation the top-level network over the
 time/logical points (e.g., chain, tree, digraph),
  ~{\it 0.3.} formulation of
 combinatorial synthesis subproblem for each time/logical point
 (i.e.,system morphological model:
  tree-like model,
 design alternatives (DAs) for leaf nodes,
 criteria for assessment of the DAs,
 estimates for DAs and their compatibility).

  {\it Phase 1.} Hierarchical system design for each time/logical point
   (combinatorial synthesis on the basis of HMMD or multiple choice problem)
  to get a set of Pareto-efficient solutions.

  {\it Phase 2.} Design of a system trajectory as a combination of
  results (i.e., solutions for the time/logical points)
  obtained at the previous phase.
  Here combinatorial synthesis
  (i.e., HMMD or multiple choice problem) is used as well.
 The resultant system trajectory can be as follows:
 (a) chain of solutions for time points
 (i.e., series or parallel-series trajectory),
 (b) tree of solutions
 (i.e., a system solution for each tree node),
 and
 (c) digraph of the system solutions
 (i.e., a system solution for each digraph node).

\begin{center}
\begin{picture}(105,63)
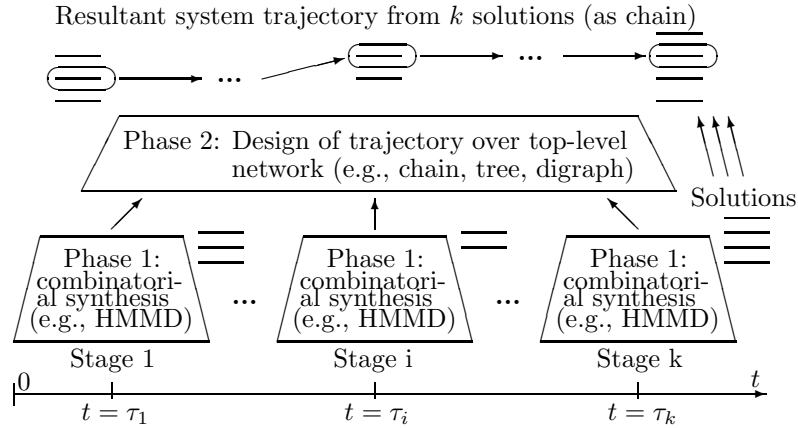
\put(17.5,00){\makebox(0,0)[bl]{Fig. 2. Illustration for
 multistage design}}

\put(0.5,9.5){\makebox(0,8)[bl]{\(0\)}}
\put(00,7.5){\line(0,1){3}}

\put(00,9){\vector(1,0){100}}
\put(98,10){\makebox(0,8)[bl]{\(t\)}}

\put(09,5){\makebox(0,8)[bl]{\(t=\tau_{1}\)}}
\put(44,5){\makebox(0,8)[bl]{\(t=\tau_{i}\)}}
\put(79,5){\makebox(0,8)[bl]{\(t=\tau_{k}\)}}

\put(13,8.5){\line(0,1){2}} \put(48,8.5){\line(0,1){2}}
\put(83,8.5){\line(0,1){2}}


\put(0,16){\line(1,0){26}}

\put(0,16){\line(1,4){3.5}} \put(26,16){\line(-1,4){3.5}}

\put(3.5,30){\line(1,0){19}}

\put(13,31){\vector(1,1){4}}

\put(6.6,26){\makebox(0,0)[bl]{Phase 1:}}
\put(3,23){\makebox(0,0)[bl]{combinatori-}}
\put(3,20){\makebox(0,0)[bl]{al synthesis }}
\put(2,17){\makebox(0,0)[bl]{(e.g., HMMD)}}

\put(24.5,30.5){\line(1,0){6}} \put(24.5,28.5){\line(1,0){6}}
\put(24.5,26.5){\line(1,0){6}}

\put(07.5,12){\makebox(0,0)[bl]{Stage 1}}


\put(29,21){\makebox(0,0)[bl]{{\bf ...}}}


\put(35,16){\line(1,0){26}}

\put(35,16){\line(1,4){3.5}} \put(61,16){\line(-1,4){3.5}}

\put(38.5,30){\line(1,0){19}}

\put(48,31){\vector(0,1){4}}

\put(41.6,26){\makebox(0,0)[bl]{Phase 1:}}
\put(38,23){\makebox(0,0)[bl]{combinatori-}}
\put(38,20){\makebox(0,0)[bl]{al synthesis }}
\put(37,17){\makebox(0,0)[bl]{(e.g., HMMD)}}

\put(59.5,30.5){\line(1,0){6}} \put(59.5,28.5){\line(1,0){6}}

\put(42.5,12){\makebox(0,0)[bl]{Stage i}}


\put(64,21){\makebox(0,0)[bl]{{\bf ...}}}


\put(70,16){\line(1,0){26}}

\put(70,16){\line(1,4){3.5}} \put(96,16){\line(-1,4){3.5}}

\put(73.5,30){\line(1,0){19}}

\put(83,31){\vector(-1,1){4}}

\put(76.6,26){\makebox(0,0)[bl]{Phase 1:}}
\put(73,23){\makebox(0,0)[bl]{combinatori-}}
\put(73,20){\makebox(0,0)[bl]{al synthesis }}
\put(72,17){\makebox(0,0)[bl]{(e.g., HMMD)}}

\put(94.5,32.5){\line(1,0){6}} \put(94.5,30.5){\line(1,0){6}}
\put(94.5,28.5){\line(1,0){6}} \put(94.5,26.5){\line(1,0){6}}


\put(93,38){\vector(-1,4){2}} \put(95,38){\vector(-1,4){2}}
\put(97,38){\vector(-1,4){2}}
\put(90,34){\makebox(0,0)[bl]{Solutions}}

\put(77.5,12){\makebox(0,0)[bl]{Stage k}}


\put(09,36){\line(1,0){79}}

\put(09,36){\line(1,2){5}} \put(88,36){\line(-1,2){5}}

\put(14,46){\line(1,0){69}}

\put(14.5,41){\makebox(0,0)[bl]{Phase 2: Design of trajectory over
top-level }}

\put(29,37){\makebox(0,0)[bl]{network (e.g., chain, tree,
digraph)}}


\put(05.5,57.4){\makebox(0,0)[bl]{Resultant system trajectory from
  \(k\) solutions (as chain)}}


\put(09,51){\oval(9,3)}

\put(5.5,54){\line(1,0){6}} \put(5.5,51){\line(1,0){6}}
\put(5.5,48){\line(1,0){6}}


\put(14,51){\vector(1,0){11}}

\put(27,50.5){\makebox(0,0)[bl]{{\bf ...}}}

\put(33,51){\vector(4,1){11}}



\put(49,54){\oval(9,3)}

\put(45.5,54){\line(1,0){6}} \put(45.5,51){\line(1,0){6}}



\put(54,54){\vector(1,0){11}}

\put(67,53.5){\makebox(0,0)[bl]{{\bf ...}}}

\put(73,54){\vector(1,0){11}}


\put(89,54){\oval(9,3)}

\put(85.5,57){\line(1,0){6}} \put(85.5,54){\line(1,0){6}}
\put(85.5,51){\line(1,0){6}} \put(85.5,48){\line(1,0){6}}

\end{picture}
\end{center}

 In Fig. 2, the following local system solutions are depicted:~
 (a) stage \(1\): \(S^{1}_{1}, S^{1}_{2}, S^{1}_{3}\);
 (b) stage \(i\): \(S^{2}_{1}, S^{2}_{2}\);
 (c) stage \(k\): \(S^{k}_{1}, S^{k}_{2}, S^{k}_{3},  S^{k}_{4}\).
 Thus, system trajectory (chain) is:~
 \(\alpha= < S^{1}_{2} \star ... \star  S^{i}_{1} \star ... \star S^{k}_{2} > \).
 An illustrative example for three-stage trajectory design is
 presented in Fig. 3,
  an example of the system trajectory (chain) is:~
 \(\beta=< S^{\tau_{1}}_{2} \star S^{\tau_{2}}_{3} \star  S^{\tau_{3}}_{1} >\).

\begin{center}
\begin{picture}(115,48)
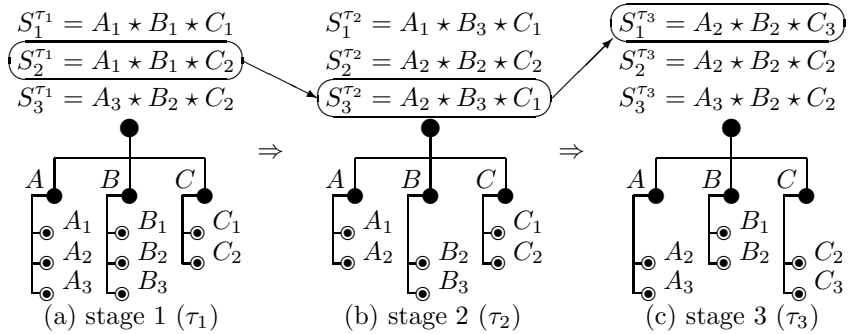
\put(10,0){\makebox(0,0)[bl]{Fig. 3. Illustrative example of
 three-stage system trajectory}}

\put(06.5,04){\makebox(0,0)[bl]{(a) stage 1 (\(\tau_{1}\))}}

\put(09,09){\makebox(0,8)[bl]{\(A_{3}\)}}
\put(09,13){\makebox(0,8)[bl]{\(A_{2}\)}}
\put(09,17){\makebox(0,8)[bl]{\(A_{1}\)}}

\put(19,09){\makebox(0,8)[bl]{\(B_{3}\)}}
\put(19,13){\makebox(0,8)[bl]{\(B_{2}\)}}
\put(19,17){\makebox(0,8)[bl]{\(B_{1}\)}}

\put(29,13){\makebox(0,8)[bl]{\(C_{2}\)}}
\put(29,17){\makebox(0,8)[bl]{\(C_{1}\)}}

\put(4,23){\makebox(0,8)[bl]{\(A\)}}
\put(14,23){\makebox(0,8)[bl]{\(B\)}}
\put(24,23){\makebox(0,8)[bl]{\(C\)}}

\put(8,22){\circle*{2}} \put(18,22){\circle*{2}}
\put(28,22){\circle*{2}}

\put(5,22){\line(1,0){02}} \put(15,22){\line(1,0){02}}
\put(25,22){\line(1,0){02}}

\put(5,22){\line(0,-1){13}} \put(15,22){\line(0,-1){13}}
\put(25,22){\line(0,-1){9}}

\put(25,13){\line(1,0){01}}
\put(25,17){\line(1,0){01}}

\put(27,13){\circle{1.8}} \put(27,13){\circle*{1}}
\put(27,17){\circle{1.8}} \put(27,17){\circle*{1}}

\put(15,09){\line(1,0){01}} \put(15,13){\line(1,0){01}}
\put(15,17){\line(1,0){01}}

\put(17,09){\circle{1.8}} \put(17,09){\circle*{1}}
\put(17,13){\circle{1.8}} \put(17,13){\circle*{1}}
\put(17,17){\circle{1.8}} \put(17,17){\circle*{1}}

\put(5,9){\line(1,0){01}} \put(5,13){\line(1,0){01}}
\put(5,17){\line(1,0){01}}

\put(7,13){\circle{1.8}} \put(7,13){\circle*{1}}
\put(7,9){\circle{1.8}} \put(7,09){\circle*{1}}
\put(7,17){\circle{1.8}} \put(7,17){\circle*{1}}

\put(8,27){\line(0,-1){04}} \put(18,27){\line(0,-1){04}}
\put(28,27){\line(0,-1){04}}

\put(8,27){\line(1,0){20}}

\put(18,27){\line(0,1){04}} \put(18,31){\circle*{2.5}}

\put(03,43){\makebox(0,8)[bl]{\(S^{\tau_{1}}_{1} = A_{1}\star
B_{1}\star C_{1}\)}}

\put(17.5,40){\oval(31,5)}

\put(33,40){\vector(2,-1){10}}

\put(03,38){\makebox(0,8)[bl]{\(S^{\tau_{1}}_{2} = A_{1}\star
B_{1}\star C_{2}\)}}

\put(03,33){\makebox(0,8)[bl]{\(S^{\tau_{1}}_{3} = A_{3}\star
B_{2}\star C_{2}\)}}

\put(35,27){\makebox(0,8)[bl]{\(\Rightarrow \)}}


\put(46.5,04){\makebox(0,0)[bl]{(b) stage 2 (\(\tau_{2}\))}}

\put(49,13){\makebox(0,8)[bl]{\(A_{2}\)}}
\put(49,17){\makebox(0,8)[bl]{\(A_{1}\)}}

\put(59,09){\makebox(0,8)[bl]{\(B_{3}\)}}
\put(59,13){\makebox(0,8)[bl]{\(B_{2}\)}}

\put(69,13){\makebox(0,8)[bl]{\(C_{2}\)}}
\put(69,17){\makebox(0,8)[bl]{\(C_{1}\)}}

\put(44,23){\makebox(0,8)[bl]{\(A\)}}
\put(54,23){\makebox(0,8)[bl]{\(B\)}}
\put(64,23){\makebox(0,8)[bl]{\(C\)}}

\put(48,22){\circle*{2}} \put(58,22){\circle*{2}}
\put(68,22){\circle*{2}}

\put(45,22){\line(1,0){02}} \put(55,22){\line(1,0){02}}
\put(65,22){\line(1,0){02}}

\put(45,22){\line(0,-1){09}} \put(55,22){\line(0,-1){13}}
\put(65,22){\line(0,-1){09}}


\put(65,17){\line(1,0){01}} \put(65,13){\line(1,0){01}}


\put(67,17){\circle{1.8}} \put(67,17){\circle*{1}}
\put(67,13){\circle{1.8}} \put(67,13){\circle*{1}}

\put(55,09){\line(1,0){01}}

\put(55,13){\line(1,0){01}}


\put(57,09){\circle{1.8}} \put(57,09){\circle*{1}}


\put(57,13){\circle{1.8}} \put(57,13){\circle*{1}}


\put(45,13){\line(1,0){01}} \put(45,17){\line(1,0){01}}

\put(47,13){\circle{1.8}} \put(47,13){\circle*{1}}


\put(47,17){\circle{1.8}} \put(47,17){\circle*{1}}

\put(48,27){\line(0,-1){04}} \put(58,27){\line(0,-1){04}}
\put(68,27){\line(0,-1){04}}

\put(48,27){\line(1,0){20}}

\put(58,27){\line(0,1){04}} \put(58,31){\circle*{2.5}}

\put(44,43){\makebox(0,8)[bl]{\(S^{\tau_{2}}_{1} = A_{1}\star
 B_{3}\star C_{1}\)}}

 \put(44,38){\makebox(0,8)[bl]{\(S^{\tau_{2}}_{2} = A_{2}\star
 B_{2}\star C_{2}\)}}

\put(58.5,35){\oval(31,5)}

\put(74.2,35){\vector(1,1){08}}

\put(44,33){\makebox(0,8)[bl]{\(S^{\tau_{2}}_{3} = A_{2}\star
 B_{3}\star C_{1}\)}}

\put(75,27){\makebox(0,8)[bl]{\(\Rightarrow \)}}


\put(86.5,04){\makebox(0,0)[bl]{(c) stage 3 (\(\tau_{3}\))}}

\put(89,09){\makebox(0,8)[bl]{\(A_{3}\)}}
\put(89,13){\makebox(0,8)[bl]{\(A_{2}\)}}

\put(99,13){\makebox(0,8)[bl]{\(B_{2}\)}}
\put(99,17){\makebox(0,8)[bl]{\(B_{1}\)}}

\put(109,09){\makebox(0,8)[bl]{\(C_{3}\)}}
\put(109,13){\makebox(0,8)[bl]{\(C_{2}\)}}

\put(84,23){\makebox(0,8)[bl]{\(A\)}}
\put(94,23){\makebox(0,8)[bl]{\(B\)}}
\put(104,23){\makebox(0,8)[bl]{\(C\)}}

\put(88,22){\circle*{2}} \put(98,22){\circle*{2}}
\put(108,22){\circle*{2}}

\put(85,22){\line(1,0){02}} \put(95,22){\line(1,0){02}}
\put(105,22){\line(1,0){02}}

\put(85,22){\line(0,-1){13}} \put(95,22){\line(0,-1){09}}
\put(105,22){\line(0,-1){13}}

\put(105,13){\line(1,0){01}} \put(105,09){\line(1,0){01}}


\put(107,13){\circle{1.8}} \put(107,13){\circle*{1}}

\put(107,09){\circle{1.8}} \put(107,09){\circle*{1}}


\put(95,13){\line(1,0){01}}


\put(95,17){\line(1,0){01}}

\put(97,17){\circle{1.8}} \put(97,17){\circle*{1}}


\put(97,13){\circle{1.8}} \put(97,13){\circle*{1}}

\put(85,9){\line(1,0){01}}


\put(85,13){\line(1,0){01}}


\put(87,09){\circle{1.8}} \put(87,09){\circle*{1}}

\put(87,13){\circle{1.8}} \put(87,13){\circle*{1}}

\put(88,27){\line(0,-1){04}} \put(98,27){\line(0,-1){04}}
\put(108,27){\line(0,-1){04}}

\put(88,27){\line(1,0){20}}

\put(98,27){\line(0,1){04}} \put(98,31){\circle*{2.5}}

\put(97.5,45){\oval(31,5)}

\put(83,43){\makebox(0,8)[bl]{\(S^{\tau_{3}}_{1} = A_{2}\star
B_{2}\star C_{3}\)}}

\put(83,38){\makebox(0,8)[bl]{\(S^{\tau_{3}}_{2} = A_{2}\star
B_{2}\star C_{2}\)}}

\put(83,33){\makebox(0,8)[bl]{\(S^{\tau_{3}}_{3} = A_{3}\star
B_{2}\star C_{2}\)}}

\end{picture}
\end{center}

 In Fig. 4,
 an illustration for tree-based system trajectory design is
 presented:

 (i) eight logical points: \(\{\mu_{0},\mu_{1},\mu_{2},\mu_{3},\mu_{4},\mu_{5},\mu_{6},\mu_{7} \}\),

 (ii) eight corresponding morphological structures:
 \(\{\Lambda^{\mu_{0}},\Lambda^{\mu_{1}},\Lambda^{\mu_{2}},\Lambda^{\mu_{3}},
 \Lambda^{\mu_{4}},\Lambda^{\mu_{5}},\Lambda^{\mu_{6}},\Lambda^{\mu_{7}} \}\).

  Here, the following solutions are depicted:~
 (a) point \(\mu_{0}\): \(S^{\mu_{0}}_{1}, S^{\mu_{0}}_{2}\);
 (b) point \(\mu_{1}\): \(S^{\mu_{1}}_{1}, S^{\mu_{1}}_{2}, S^{\mu_{1}}_{3}\);
 (c) point \(\mu_{2}\): \(S^{\mu_{2}}_{1}, S^{\mu_{2}}_{2}\);
 (d) point \(\mu_{3}\): \(S^{\mu_{3}}_{1}, S^{\mu_{3}}_{2}, S^{\mu_{3}}_{3}\);
 (e) point \(\mu_{4}\): \(S^{\mu_{4}}_{1}, S^{\mu_{4}}_{2}, S^{\mu_{4}}_{3}\);
 (f) point \(\mu_{5}\): \(S^{\mu_{5}}_{1}, S^{\mu_{5}}_{2}, S^{\mu_{5}}_{3}\);
 (h) point \(\mu_{6}\): \(S^{\mu_{6}}_{1}, S^{\mu_{6}}_{2},
 S^{\mu_{6}}_{3}\);
 (i) point \(\mu_{7}\): \(S^{\mu_{6}}_{1}, S^{\mu_{6}}_{2}\).
 Thus, system trajectory (tree) consists of the following local solutions:~
 \(\gamma= \{ S^{\mu_{0}}_{1}, S^{\mu_{1}}_{1}, S^{\mu_{2}}_{2},
 S^{\mu_{3}}_{1}, S^{\mu_{4}}_{3}, S^{\mu_{5}}_{1},  S^{\mu_{6}}_{2}, S^{\mu_{7}}_{1} \}
 \).
 Clearly, the structure of the  system trajectory \(\gamma \)
 corresponds to the initial tree (Fig. 5).

\begin{center}
\begin{picture}(55,70)

\put(11,0){\makebox(0,0)[bl]{Fig. 4. Illustration  for tree based
system trajectory}}

\put(01,05){\makebox(0,0)[bl]{(a) top-level network (tree)}}

\put(4,29){\oval(8,6)} \put(4,29){\oval(7,5)}
\put(02,28){\makebox(0,8)[bl]{\(\mu_{0}\)}}

\put(09,28){\vector(1,-1){08}}

\put(09,30){\vector(1,1){08}}


\put(22,19){\oval(8,6)} \put(22,19){\oval(7,5)}
\put(20,18){\makebox(0,8)[bl]{\(\mu_{1}\)}}

\put(27,19){\vector(1,0){08}}

\put(26,22){\vector(3,2){09}}


\put(40,19){\oval(8,6)} \put(40,19){\oval(7,5)}
\put(38,18){\makebox(0,8)[bl]{\(\mu_{2}\)}}


\put(40,29){\oval(8,6)} \put(40,29){\oval(7,5)}
\put(38,28){\makebox(0,8)[bl]{\(\mu_{3}\)}}



\put(22,39){\oval(8,6)} \put(22,39){\oval(7,5)}
\put(20,38){\makebox(0,8)[bl]{\(\mu_{4}\)}}

\put(27,39){\vector(1,0){08}}

\put(26,42){\vector(3,2){09}}

\put(24,43){\line(1,2){7}} \put(31,57){\vector(2,1){04}}


\put(40,39){\oval(8,6)} \put(40,39){\oval(7,5)}
\put(38,38){\makebox(0,8)[bl]{\(\mu_{5}\)}}


\put(40,49){\oval(8,6)} \put(40,49){\oval(7,5)}
\put(38,48){\makebox(0,8)[bl]{\(\mu_{6}\)}}


\put(40,59){\oval(8,6)} \put(40,59){\oval(7,5)}
\put(38,58){\makebox(0,8)[bl]{\(\mu_{7}\)}}

\end{picture}
%
\begin{picture}(50,80)

\put(08,05){\makebox(0,0)[bl]{(b) system trajectory}}




\put(02.5,36){\oval(5,2)}


\put(00.5,36){\line(1,0){4}} \put(00.5,34.5){\line(1,0){4}}

\put(05,33.5){\circle*{1.5}}

\put(06.5,34){\line(1,1){10}} \put(16.5,44){\vector(3,2){07}}
\put(06.5,33){\line(1,-1){10}} \put(16.5,23){\vector(3,-2){07}}

\put(00,26){\line(2,3){5}} \put(10,26){\line(-2,3){5}}
\put(00,26){\line(1,0){10}}

\put(02.2,27){\makebox(0,8)[bl]{\(\Lambda^{\mu_{0}}\)}}


\put(23.5,23.5){\oval(5,2)}

\put(21.5,23.5){\line(1,0){4}} \put(21.5,22){\line(1,0){4}}
\put(21.5,20.5){\line(1,0){4}}

\put(25,18.5){\circle*{1.5}}

\put(26.4,18.5){\vector(1,0){17.2}}

\put(26,19.5){\vector(4,3){18}}

\put(20,11){\line(2,3){5}} \put(30,11){\line(-2,3){5}}
\put(20,11){\line(1,0){10}}

\put(22.2,12){\makebox(0,8)[bl]{\(\Lambda^{\mu_{1}}\)}}


\put(42.5,21){\oval(5,2)}

\put(40.5,22.5){\line(1,0){4}} \put(40.5,21){\line(1,0){4}}

\put(45,18.5){\circle*{1.5}}

\put(40,11){\line(2,3){5}} \put(50,11){\line(-2,3){5}}
\put(40,11){\line(1,0){10}}

\put(42.2,12){\makebox(0,8)[bl]{\(\Lambda^{\mu_{2}}\)}}


\put(42.5,37.5){\oval(5,2)}

\put(40.5,37.5){\line(1,0){4}} \put(40.5,36){\line(1,0){4}}
\put(40.5,34.5){\line(1,0){4}}

\put(45,33.5){\circle*{1.5}}

\put(40,26){\line(2,3){5}} \put(50,26){\line(-2,3){5}}
\put(40,26){\line(1,0){10}}

\put(42.2,27){\makebox(0,8)[bl]{\(\Lambda^{\mu_{3}}\)}}



\put(22,50.5){\oval(5,2)}

\put(20,53.5){\line(1,0){4}} \put(20,52){\line(1,0){4}}
\put(20,50.5){\line(1,0){4}}

\put(25,48.5){\circle*{1.5}}

\put(26.4,48.5){\vector(1,0){17.2}}
\put(26,49.5){\vector(4,3){18}}

\put(25,50){\line(1,3){07.4}} \put(32.4,72.2){\vector(2,1){11.1}}

\put(20,41){\line(2,3){5}} \put(30,41){\line(-2,3){5}}
\put(20,41){\line(1,0){10}}

\put(22.2,42){\makebox(0,8)[bl]{\(\Lambda^{\mu_{4}}\)}}


\put(42.5,52.5){\oval(5,2)}

\put(40.5,52.5){\line(1,0){4}} \put(40.5,51){\line(1,0){4}}
\put(40.5,49.5){\line(1,0){4}}

\put(45,48.5){\circle*{1.5}}

\put(40,41){\line(2,3){5}} \put(50,41){\line(-2,3){5}}
\put(40,41){\line(1,0){10}}

\put(42.2,42){\makebox(0,8)[bl]{\(\Lambda^{\mu_{5}}\)}}


\put(39.5,64){\oval(5,2)}

\put(37.5,65.5){\line(1,0){4}} \put(37.5,64){\line(1,0){4}}
\put(37.5,62.5){\line(1,0){4}}

\put(45,63.5){\circle*{1.5}}

\put(40,56){\line(2,3){5}} \put(50,56){\line(-2,3){5}}
\put(40,56){\line(1,0){10}}

\put(42.2,57){\makebox(0,8)[bl]{\(\Lambda^{\mu_{6}}\)}}


\put(39.5,79){\oval(5,2)}


\put(37.5,79){\line(1,0){4}} \put(37.5,77.5){\line(1,0){4}}

\put(45,78.5){\circle*{1.5}}

\put(40,71){\line(2,3){5}} \put(50,71){\line(-2,3){5}}
\put(40,71){\line(1,0){10}}

\put(42.2,72){\makebox(0,8)[bl]{\(\Lambda^{\mu_{7}}\)}}

\end{picture}
\end{center}

\begin{center}
\begin{picture}(65,54)

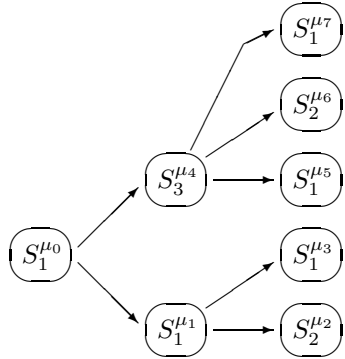
\put(00,00){\makebox(0,0)[bl]{Fig. 5. Example of tree-like
 trajectory}}

\put(4,19){\oval(8,7)}
\put(01.5,17){\makebox(0,8)[bl]{\(S^{\mu_{0}}_{1}\)}}

\put(09,18){\vector(1,-1){08}}

\put(09,20){\vector(1,1){08}}


\put(22,09){\oval(8,7)}
\put(19.5,07){\makebox(0,8)[bl]{\(S^{\mu_{1}}_{1}\)}}

\put(27,09){\vector(1,0){08}}

\put(26,12){\vector(3,2){09}}


\put(40,09){\oval(8,7)}
\put(37.5,07){\makebox(0,8)[bl]{\(S^{\mu_{2}}_{2}\)}}


\put(40,19){\oval(8,7)}
\put(37.5,17){\makebox(0,8)[bl]{\(S^{\mu_{3}}_{1}\)}}



\put(22,29){\oval(8,7)}
\put(19.5,27){\makebox(0,8)[bl]{\(S^{\mu_{4}}_{3}\)}}

\put(27,29){\vector(1,0){08}}

\put(26,32){\vector(3,2){09}}

\put(24,33){\line(1,2){7}} \put(31,47){\vector(2,1){04}}


\put(40,29){\oval(8,7)}
\put(37.5,27){\makebox(0,8)[bl]{\(S^{\mu_{5}}_{1}\)}}


\put(40,39){\oval(8,7)}
\put(37.5,37){\makebox(0,8)[bl]{\(S^{\mu_{6}}_{2}\)}}


\put(40,49){\oval(8,7)}
\put(37.5,47){\makebox(0,8)[bl]{\(S^{\mu_{7}}_{1}\)}}

\end{picture}
\end{center}

 In the case of tree-based system trajectories,
 it is reasonable to use additional nodes (as 'analysis/decision' points)
 for an analysis of the implementation results
 and selection of the next direction.
 Fig. 6 depicts an example of this kind of the extended tree-like network
 with corresponding additional  'analysis/decision' points:  \(a_{0}\), \(a_{1}\), \(a_{4}\).
 Here,
 the resultant system trajectory is a chain
 (from the root to a leaf node),
 for example (for Fig. 6):~
  \(\gamma' =
   < S^{\mu_{0}} \star  S^{\mu_{4}} \star S^{\mu_{5}} > \).

\begin{center}
\begin{picture}(76,53)

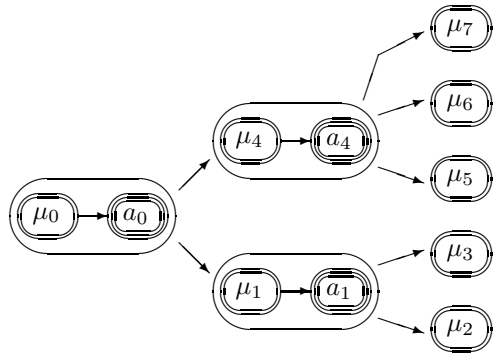
\put(00,00){\makebox(0,0)[bl]{Fig. 6. Tree-like network with
 'analysis' nodes}}


\put(11,24){\oval(22,10)}

\put(05,24){\oval(8,6)} \put(05,24){\oval(7,5)}
\put(03,23){\makebox(0,8)[bl]{\(\mu_{0}\)}}

\put(09,24){\vector(1,0){4}}


\put(17,24){\oval(8,6)} \put(17,24){\oval(7,5)}
\put(17,24){\oval(6,4)}

\put(15,23){\makebox(0,8)[bl]{\(a_{0}\)}}

\put(22.5,27.5){\vector(1,1){4}} \put(22.5,20.5){\vector(1,-1){4}}


\put(38,14){\oval(22,10)}

\put(32,14){\oval(8,6)} \put(32,14){\oval(7,5)}
\put(30,13){\makebox(0,8)[bl]{\(\mu_{1}\)}}

\put(36,14){\vector(1,0){4}}


\put(44,14){\oval(8,6)} \put(44,14){\oval(7,5)}
\put(44,14){\oval(6,4)}

\put(42,13){\makebox(0,8)[bl]{\(a_{1}\)}}

\put(49,17.5){\vector(3,1){6}} \put(49,10.5){\vector(3,-1){6}}


\put(60,09){\oval(8,6)} \put(60,09){\oval(7,5)}
\put(58,08){\makebox(0,8)[bl]{\(\mu_{2}\)}}


\put(60,19){\oval(8,6)} \put(60,19){\oval(7,5)}
\put(58,18){\makebox(0,8)[bl]{\(\mu_{3}\)}}



\put(38,34){\oval(22,10)}

\put(32,34){\oval(8,6)} \put(32,34){\oval(7,5)}
\put(30,33){\makebox(0,8)[bl]{\(\mu_{4}\)}}

\put(36,34){\vector(1,0){4}}


\put(44,34){\oval(8,6)} \put(44,34){\oval(7,5)}
\put(44,34){\oval(6,4)}

\put(42,33){\makebox(0,8)[bl]{\(a_{4}\)}}

\put(49,37.5){\vector(3,1){6}} \put(49,30.5){\vector(3,-1){6}}

\put(47,39.5){\line(1,3){2}} \put(49,45.5){\vector(2,1){6}}


\put(60,29){\oval(8,6)} \put(60,29){\oval(7,5)}
\put(58,28){\makebox(0,8)[bl]{\(\mu_{5}\)}}


\put(60,39){\oval(8,6)} \put(60,39){\oval(7,5)}
\put(58,38){\makebox(0,8)[bl]{\(\mu_{6}\)}}


\put(60,49){\oval(8,6)} \put(60,49){\oval(7,5)}
\put(58,48){\makebox(0,8)[bl]{\(\mu_{7}\)}}


\end{picture}
\end{center}

 Fig. 7 depicts an example of digraph with corresponding
 system trajectories.
 Here,
 the following situations (problems) can be examined:
 ~(i) design of a route (e.g., series as a chain-route, series-parallel route, tree-like route)
 on the basis of the initial digraph;
 ~(ii) design of a spanning tree for the initial digraph and
 study of the previous problem for tree-like network
 (including usage of the additional 'analysis' nodes).
 Illustrative examples of routes  are (Fig. 7):
 (a) chain-route (series):~
 \( < S_{2}^{\mu_{4}} \star  S_{3}^{\mu_{7}} \star  S_{2}^{\mu_{8}} > \),
  (b) tree-like route on the basis of seven  solutions:~
  \(\{
  S_{2}^{\mu_{4}},S_{1}^{\mu_{5}},S_{1}^{\mu_{6}},S_{2}^{\mu_{8}},
  S_{1}^{\mu_{1}},S_{2}^{\mu_{2}},S_{1}^{\mu_{3}} \}\), and
 (c) series-parallel route on the basis of four solutions:~
 \(\{ S_{2}^{\mu_{4}},S_{1}^{\mu_{5}},S_{3}^{\mu_{7}},S_{2}^{\mu_{8}} \}\),

\begin{center}
\begin{picture}(55,64)

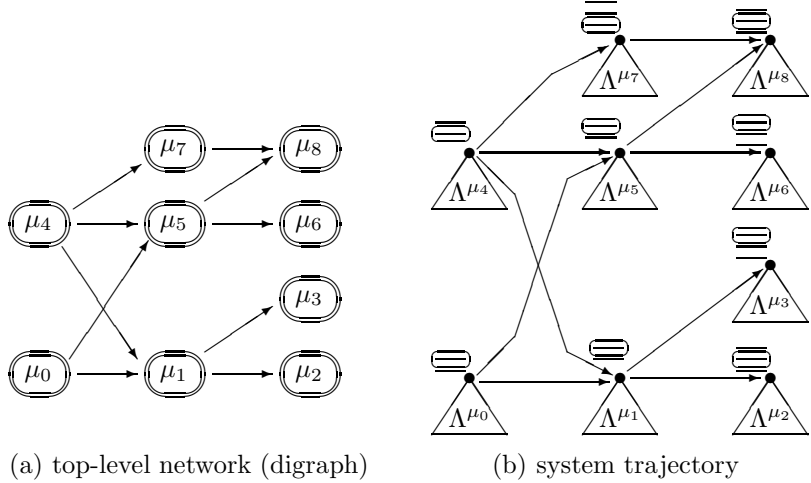
\put(07,0){\makebox(0,0)[bl]{Fig. 7. Illustration  for digraph
based system trajectory}}

\put(00,05){\makebox(0,0)[bl]{(a) top-level network (digraph)}}

\put(4,19){\oval(8,6)} \put(4,19){\oval(7,5)}
\put(02,18){\makebox(0,8)[bl]{\(\mu_{0}\)}}

\put(09,19){\vector(1,0){08}}

\put(08,21){\vector(2,3){10.5}}


\put(4,39){\oval(8,6)} \put(4,39){\oval(7,5)}
\put(02,38){\makebox(0,8)[bl]{\(\mu_{4}\)}}

\put(09,39){\vector(1,0){08}}


\put(07,36){\vector(2,-3){10}}


\put(08.5,41){\vector(3,2){09}}


\put(22,19){\oval(8,6)} \put(22,19){\oval(7,5)}
\put(20,18){\makebox(0,8)[bl]{\(\mu_{1}\)}}

\put(27,19){\vector(1,0){08}}

\put(26,22){\vector(3,2){09}}




\put(40,19){\oval(8,6)} \put(40,19){\oval(7,5)}
\put(38,18){\makebox(0,8)[bl]{\(\mu_{2}\)}}


\put(40,29){\oval(8,6)} \put(40,29){\oval(7,5)}
\put(38,28){\makebox(0,8)[bl]{\(\mu_{3}\)}}



\put(22,39){\oval(8,6)} \put(22,39){\oval(7,5)}
\put(20,38){\makebox(0,8)[bl]{\(\mu_{5}\)}}

\put(27,39){\vector(1,0){08}}

\put(26,42){\vector(3,2){09}}


\put(22,49){\oval(8,6)} \put(22,49){\oval(7,5)}
\put(20,48){\makebox(0,8)[bl]{\(\mu_{7}\)}}

\put(27,49){\vector(1,0){08}}





\put(40,39){\oval(8,6)} \put(40,39){\oval(7,5)}
\put(38,38){\makebox(0,8)[bl]{\(\mu_{6}\)}}


\put(40,49){\oval(8,6)} \put(40,49){\oval(7,5)}
\put(38,48){\makebox(0,8)[bl]{\(\mu_{8}\)}}

\end{picture}
%
\begin{picture}(50,69)

\put(08,05){\makebox(0,0)[bl]{(b) system trajectory}}


\put(06,19){\line(1,1){5}} \put(11,24){\line(1,3){7}}
\put(018,45){\vector(2,1){6}}


\put(02.5,21){\oval(5,2)}


\put(00.5,21){\line(1,0){4}} \put(00.5,19.5){\line(1,0){4}}

\put(05,18.5){\circle*{1.5}}

\put(06.4,18){\vector(1,0){17.2}}

\put(00,11){\line(2,3){5}} \put(10,11){\line(-2,3){5}}
\put(00,11){\line(1,0){10}}

\put(02.2,12){\makebox(0,8)[bl]{\(\Lambda^{\mu_{0}}\)}}


\put(06,48){\line(1,-1){5}} \put(11,43){\line(1,-3){7}}
\put(18,22){\vector(2,-1){6}}

\put(06,49){\line(1,1){10}} \put(16,59){\vector(2,1){7.5}}



\put(02.5,51){\oval(5,2)}

\put(00.5,52.5){\line(1,0){4}} \put(00.5,51){\line(1,0){4}}

\put(05,48.5){\circle*{1.5}}

\put(06.4,48.5){\vector(1,0){17.2}}

\put(00,41){\line(2,3){5}} \put(10,41){\line(-2,3){5}}
\put(00,41){\line(1,0){10}}

\put(02.2,42){\makebox(0,8)[bl]{\(\Lambda^{\mu_{4}}\)}}


\put(23.5,22.5){\oval(5,2)}

\put(21.5,22.5){\line(1,0){4}} \put(21.5,21){\line(1,0){4}}


\put(25,18.5){\circle*{1.5}}

\put(26.4,18.5){\vector(1,0){17.2}}

\put(26,19.5){\vector(4,3){18}}

\put(20,11){\line(2,3){5}} \put(30,11){\line(-2,3){5}}
\put(20,11){\line(1,0){10}}

\put(22.2,12){\makebox(0,8)[bl]{\(\Lambda^{\mu_{1}}\)}}


\put(42.5,21){\oval(5,2)}

\put(40.5,22.5){\line(1,0){4}} \put(40.5,21){\line(1,0){4}}
\put(40.5,19.5){\line(1,0){4}}

\put(45,18.5){\circle*{1.5}}  \put(40,11){\line(2,3){5}}
\put(50,11){\line(-2,3){5}}
\put(40,11){\line(1,0){10}}

\put(42.2,12){\makebox(0,8)[bl]{\(\Lambda^{\mu_{2}}\)}}


\put(42.5,37.5){\oval(5,2)}

\put(40.5,37.5){\line(1,0){4}} \put(40.5,36){\line(1,0){4}}
\put(40.5,34.5){\line(1,0){4}}

\put(45,33.5){\circle*{1.5}}

\put(40,26){\line(2,3){5}} \put(50,26){\line(-2,3){5}}
\put(40,26){\line(1,0){10}}

\put(42.2,27){\makebox(0,8)[bl]{\(\Lambda^{\mu_{3}}\)}}



\put(22.5,52){\oval(5,2)}


\put(20.5,52){\line(1,0){4}} \put(20.5,50.5){\line(1,0){4}}

\put(25,48.5){\circle*{1.5}}

\put(26.4,48.5){\vector(1,0){17.2}}

\put(26,49.5){\vector(4,3){18}}

\put(20,41){\line(2,3){5}} \put(30,41){\line(-2,3){5}}
\put(20,41){\line(1,0){10}}

\put(22.2,42){\makebox(0,8)[bl]{\(\Lambda^{\mu_{5}}\)}}


\put(22.5,65.5){\oval(5,2)}

\put(20.5,68.5){\line(1,0){4}} \put(20.5,67){\line(1,0){4}}
\put(20.5,65.5){\line(1,0){4}}

\put(25,63.5){\circle*{1.5}}

\put(26.4,63.5){\vector(1,0){17.2}}


\put(20,56){\line(2,3){5}} \put(30,56){\line(-2,3){5}}
\put(20,56){\line(1,0){10}}

\put(22.2,57){\makebox(0,8)[bl]{\(\Lambda^{\mu_{7}}\)}}


\put(42.5,52.5){\oval(5,2)}

\put(40.5,52.5){\line(1,0){4}} \put(40.5,51){\line(1,0){4}}
\put(40.5,49.5){\line(1,0){4}}

\put(45,48.5){\circle*{1.5}}

\put(40,41){\line(2,3){5}} \put(50,41){\line(-2,3){5}}
\put(40,41){\line(1,0){10}}

\put(42.2,42){\makebox(0,8)[bl]{\(\Lambda^{\mu_{6}}\)}}


\put(42.5,66){\oval(5,2)}

\put(40.5,67.5){\line(1,0){4}} \put(40.5,66){\line(1,0){4}}
\put(40.5,64.5){\line(1,0){4}}

\put(45,63.5){\circle*{1.5}}

\put(40,56){\line(2,3){5}} \put(50,56){\line(-2,3){5}}
\put(40,56){\line(1,0){10}}

\put(42.2,57){\makebox(0,8)[bl]{\(\Lambda^{\mu_{8}}\)}}

\end{picture}
\end{center}

 The initial top-level network (Fig. 7a)
 can be approximate by a simple spanning structure:~
 (a) chain (Fig. 8a),
 (b) spanning (approximate) tree (Fig. 8b),
 (c) spanning simplified network (Fig. 8c).

\begin{center}
\begin{picture}(37,48.5)

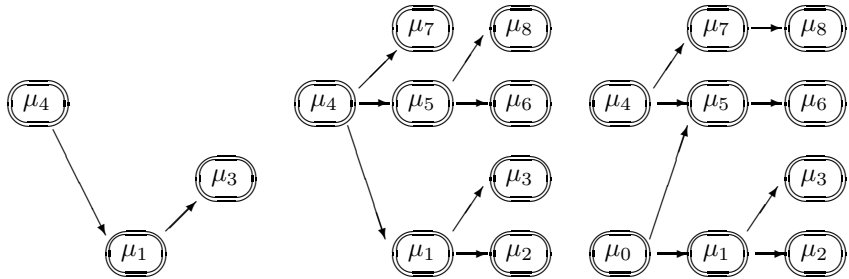
\put(01.6,00){\makebox(0,0)[bl]{Fig. 8. Spanning (approximate)
 structures for basic top-level network}}

\put(00,05){\makebox(0,0)[bl]{(a) chain
 \(<\mu_{4},\mu_{1},\mu_{3} >\)}}


\put(4,34){\oval(8,6)} \put(4,34){\oval(7,5)}
\put(02,33){\makebox(0,8)[bl]{\(\mu_{4}\)}}

\put(06,30.5){\vector(1,-2){07}}


\put(17,14){\oval(8,6)} \put(17,14){\oval(7,5)}
\put(15,13){\makebox(0,8)[bl]{\(\mu_{1}\)}}


\put(21,17){\vector(1,1){04}}


\put(29,24){\oval(8,6)} \put(29,24){\oval(7,5)}
\put(27,23){\makebox(0,8)[bl]{\(\mu_{3}\)}}

\end{picture}
%
\begin{picture}(38,48)


\put(00,05){\makebox(0,0)[bl]{(b) approximate tree}}


\put(4,34){\oval(8,6)} \put(4,34){\oval(7,5)}
\put(02,33){\makebox(0,8)[bl]{\(\mu_{4}\)}}

\put(08.5,34){\vector(1,0){04}}

\put(07,31){\vector(1,-3){05}}

\put(08.5,36){\vector(1,1){05}}


\put(17,14){\oval(8,6)} \put(17,14){\oval(7,5)}
\put(15,13){\makebox(0,8)[bl]{\(\mu_{1}\)}}

\put(21.5,14){\vector(1,0){04}}

\put(21,17){\vector(2,3){04}}


\put(30,14){\oval(8,6)} \put(30,14){\oval(7,5)}
\put(28,13){\makebox(0,8)[bl]{\(\mu_{2}\)}}


\put(30,24){\oval(8,6)} \put(30,24){\oval(7,5)}
\put(28,23){\makebox(0,8)[bl]{\(\mu_{3}\)}}



\put(17,34){\oval(8,6)} \put(17,34){\oval(7,5)}
\put(15,33){\makebox(0,8)[bl]{\(\mu_{5}\)}}

\put(21.5,34){\vector(1,0){04}}

\put(21,37){\vector(2,3){04}}


\put(17,44){\oval(8,6)} \put(17,44){\oval(7,5)}
\put(15,43){\makebox(0,8)[bl]{\(\mu_{7}\)}}


\put(30,34){\oval(8,6)} \put(30,34){\oval(7,5)}
\put(28,33){\makebox(0,8)[bl]{\(\mu_{6}\)}}


\put(30,44){\oval(8,6)} \put(30,44){\oval(7,5)}
\put(28,43){\makebox(0,8)[bl]{\(\mu_{8}\)}}

\end{picture}
%
\begin{picture}(34,48)


\put(00,05){\makebox(0,0)[bl]{(c) simplified network }}

\put(4,14){\oval(8,6)} \put(4,14){\oval(7,5)}
\put(02,13){\makebox(0,8)[bl]{\(\mu_{0}\)}}

\put(09,14){\vector(1,0){04}}

\put(08,16){\vector(1,3){05}}


\put(4,34){\oval(8,6)} \put(4,34){\oval(7,5)}
\put(02,33){\makebox(0,8)[bl]{\(\mu_{4}\)}}

\put(09,34){\vector(1,0){04}}

\put(08.5,36){\vector(2,3){04}}


\put(17,14){\oval(8,6)} \put(17,14){\oval(7,5)}
\put(15,13){\makebox(0,8)[bl]{\(\mu_{1}\)}}

\put(22,14){\vector(1,0){04}}

\put(21,17){\vector(2,3){04}}


\put(30,14){\oval(8,6)} \put(30,14){\oval(7,5)}
\put(28,13){\makebox(0,8)[bl]{\(\mu_{2}\)}}


\put(30,24){\oval(8,6)} \put(30,24){\oval(7,5)}
\put(28,23){\makebox(0,8)[bl]{\(\mu_{3}\)}}



\put(17,34){\oval(8,6)} \put(17,34){\oval(7,5)}
\put(15,33){\makebox(0,8)[bl]{\(\mu_{5}\)}}

\put(21.5,34){\vector(1,0){04}}


\put(17,44){\oval(8,6)} \put(17,44){\oval(7,5)}
\put(15,43){\makebox(0,8)[bl]{\(\mu_{7}\)}}

\put(21.5,44){\vector(1,0){04}}


\put(30,34){\oval(8,6)} \put(30,34){\oval(7,5)}
\put(28,33){\makebox(0,8)[bl]{\(\mu_{6}\)}}


\put(30,44){\oval(8,6)} \put(30,44){\oval(7,5)}
\put(28,43){\makebox(0,8)[bl]{\(\mu_{8}\)}}

\end{picture}
\end{center}

\section{Applied Illustrative Examples}


\subsection{Four-Stage Trajectory for Start-Up Team}

 Here, the basic version of HMMD approach is used
  (e.g., \cite{lev98,lev06,lev12morph}).
 The example is an illustrative one
 (expert judgment).
 A general examined tree-like structure for the  start-up team is:

~~

 {\bf 0.} Tree-like structure  ~\(S = L \star R \star E \star M  \):

 {\it 1.} Creator or leader of the project ~\(L\):
 none \(L_{0}\),
 creator (part-time participation, consulting) \(L_{1}\),
 creator (full-time participation)  \(L_{2}\).

 {\it 2.} Researcher  ~\(R\):
 none \(R_{0}\),
 researcher (part time participation, consulting)
   \(R_{1}\),
  researcher (full-time participation)  \(R_{2}\),
 two researchers \(R_{3} = R_{1} \&R_{2}\).

 {\it 3.} Engineer ~\(E\):
 none \(E_{0}\),
 researcher (part time participation, consulting)
   \(E_{1}\),
 researcher (full-time participation)  \(E_{2}\),
 two engineers \(E_{3} = E_{1} \& E_{2}\).

 {\it 4.} Manager ~\(M\):
 none \(M_{0}\),
 manager (part time participation, consulting)
   \(M_{1}\),
 manager (full-time participation)  \(M_{2}\).


 This structure is analyzed for the following four stages:~

~~

  {\it Stage 0} (\(t=\tau_{0}\))
  (Fig. 9, priorities of DAs are shown in parentheses;
  Table 2):~
 creation of the basic idea for a new product/system,
  preparation of the project proposal.

 {\it Stage 1} (\(t=\tau_{1}\)) (Fig. 10,
 priorities of DAs are shown in parentheses;
 Table 3):~
 design of a system prototype,
 preparation of research materials as papers, presentation at
 conference,
 preparation of a patent,
 searching for investors.

 {\it Stage 2} (\(t=\tau_{2}\)) (Fig. 11,
 priorities of DAs are shown in parentheses;
 Table 4):~
 design of a preliminary system version,
 analysis of the markets,
 preparation of business plan(s),
 searching for investors.

 {\it Stage 3} (\(t=\tau_{3}\)) (Fig. 12,
 priorities of DAs are shown in parentheses;
 Table 5):~
 design of a system version,
 searching for applied domains,
 marketing,
 customization.

\begin{center}
\begin{picture}(55,39)

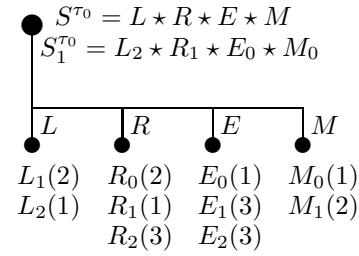
\put(00,00){\makebox(0,0)[bl] {Fig. 9. Structure of team
(\(\tau_{0}\))}}

\put(1,13){\makebox(0,8)[bl]{\(L_{1}(2)\)}}
\put(1,09){\makebox(0,8)[bl]{\(L_{2}(1)\)}}

\put(13,13){\makebox(0,8)[bl]{\(R_{0}(2)\)}}
\put(13,09){\makebox(0,8)[bl]{\(R_{1}(1)\)}}
\put(13,05){\makebox(0,8)[bl]{\(R_{2}(3)\)}}

\put(25,13){\makebox(0,8)[bl]{\(E_{0}(1)\)}}
\put(25,09){\makebox(0,8)[bl]{\(E_{1}(3)\)}}
\put(25,05){\makebox(0,8)[bl]{\(E_{2}(3)\)}}

\put(37,13){\makebox(0,8)[bl]{\(M_{0}(1)\)}}
\put(37,09){\makebox(0,8)[bl]{\(M_{1}(2)\)}}


\put(03,19){\circle*{2}} \put(15,19){\circle*{2}}
\put(27,19){\circle*{2}} \put(39,19){\circle*{2}}

\put(03,24){\line(0,-1){04}} \put(15,24){\line(0,-1){04}}
\put(27,24){\line(0,-1){04}} \put(39,24){\line(0,-1){04}}


\put(03,24){\line(1,0){36}}

\put(04,20.5){\makebox(0,8)[bl]{\(L\) }}
\put(16,20.5){\makebox(0,8)[bl]{\(R\) }}
\put(28,20.5){\makebox(0,8)[bl]{\(E\) }}
\put(40,20.5){\makebox(0,8)[bl]{\(M\) }}


\put(03,24){\line(0,1){10}} \put(03,35){\circle*{2.7}}

\put(06,35){\makebox(0,8)[bl] {\(S^{\tau_{0}} =
 L \star R \star E \star M \)}}

\put(04,30){\makebox(0,8)[bl] {\(S^{\tau_{0}}_{1} =
 L_{2} \star R_{1} \star E_{0} \star M_{0}
 \)}}



\end{picture}
%
\begin{picture}(54,46)

\put(06,41.5){\makebox(0,0)[bl]{Table 2. Compatibility
(\(\tau_{0}\))}}

\put(00,0){\line(1,0){54}} \put(00,34){\line(1,0){54}}
\put(00,40){\line(1,0){54}}

\put(00,0){\line(0,1){40}} \put(06,0){\line(0,1){40}}
\put(54,0){\line(0,1){40}}

\put(01,30){\makebox(0,0)[bl]{\(L_{1}\)}}
\put(01,26){\makebox(0,0)[bl]{\(L_{2}\)}}

\put(01,22){\makebox(0,0)[bl]{\(R_{0}\)}}
\put(01,18){\makebox(0,0)[bl]{\(R_{1}\)}}
\put(01,14){\makebox(0,0)[bl]{\(R_{2}\)}}

\put(01,10){\makebox(0,0)[bl]{\(E_{0}\)}}
\put(01,06){\makebox(0,0)[bl]{\(E_{1}\)}}
\put(01,02){\makebox(0,0)[bl]{\(E_{2}\)}}

\put(12,34){\line(0,1){6}} \put(18,34){\line(0,1){6}}
\put(24,34){\line(0,1){6}} \put(30,34){\line(0,1){6}}
\put(36,34){\line(0,1){6}} \put(42,34){\line(0,1){6}}
\put(48,34){\line(0,1){6}}

\put(07,36){\makebox(0,0)[bl]{\(R_{0}\)}}
\put(13,36){\makebox(0,0)[bl]{\(R_{1}\)}}
\put(19,36){\makebox(0,0)[bl]{\(R_{2}\)}}
\put(25,36){\makebox(0,0)[bl]{\(E_{0}\)}}
\put(31,36){\makebox(0,0)[bl]{\(E_{1}\)}}
\put(37,36){\makebox(0,0)[bl]{\(E_{2}\)}}
\put(42.6,36){\makebox(0,0)[bl]{\(M_{0}\)}}
\put(48.6,36){\makebox(0,0)[bl]{\(M_{1}\)}}


\put(08,30){\makebox(0,0)[bl]{\(1\)}}
\put(14,30){\makebox(0,0)[bl]{\(2\)}}
\put(20,30){\makebox(0,0)[bl]{\(1\)}}
\put(26,30){\makebox(0,0)[bl]{\(3\)}}
\put(32,30){\makebox(0,0)[bl]{\(3\)}}
\put(38,30){\makebox(0,0)[bl]{\(2\)}}
\put(44,30){\makebox(0,0)[bl]{\(3\)}}
\put(50,30){\makebox(0,0)[bl]{\(2\)}}

\put(08,26){\makebox(0,0)[bl]{\(1\)}}
\put(14,26){\makebox(0,0)[bl]{\(3\)}}
\put(20,26){\makebox(0,0)[bl]{\(2\)}}
\put(26,26){\makebox(0,0)[bl]{\(3\)}}
\put(32,26){\makebox(0,0)[bl]{\(3\)}}
\put(38,26){\makebox(0,0)[bl]{\(2\)}}
\put(44,26){\makebox(0,0)[bl]{\(3\)}}
\put(50,26){\makebox(0,0)[bl]{\(2\)}}

\put(26,22){\makebox(0,0)[bl]{\(3\)}}
\put(32,22){\makebox(0,0)[bl]{\(3\)}}
\put(38,22){\makebox(0,0)[bl]{\(2\)}}
\put(44,22){\makebox(0,0)[bl]{\(3\)}}
\put(50,22){\makebox(0,0)[bl]{\(1\)}}

\put(26,18){\makebox(0,0)[bl]{\(3\)}}
\put(32,18){\makebox(0,0)[bl]{\(3\)}}
\put(38,18){\makebox(0,0)[bl]{\(1\)}}
\put(44,18){\makebox(0,0)[bl]{\(3\)}}
\put(50,18){\makebox(0,0)[bl]{\(3\)}}

\put(26,14){\makebox(0,0)[bl]{\(3\)}}
\put(32,14){\makebox(0,0)[bl]{\(3\)}}
\put(38,14){\makebox(0,0)[bl]{\(1\)}}
\put(44,14){\makebox(0,0)[bl]{\(3\)}}
\put(50,14){\makebox(0,0)[bl]{\(2\)}}

\put(44,10){\makebox(0,0)[bl]{\(3\)}}
\put(50,10){\makebox(0,0)[bl]{\(3\)}}

\put(44,06){\makebox(0,0)[bl]{\(3\)}}
\put(50,06){\makebox(0,0)[bl]{\(3\)}}

\put(44,02){\makebox(0,0)[bl]{\(3\)}}
\put(50,02){\makebox(0,0)[bl]{\(2\)}}

\end{picture}
\end{center}

 The resultant composite Pareto-efficient DA is the following (Fig. 13):

 \(S^{\tau_{0}}_{1} = L_{2} \star R_{1} \star E_{0} \star M_{0} \),
 \(N(S^{\tau_{0}}_{1})= (3;4,0,0)\).

\begin{center}
\begin{picture}(55,39)

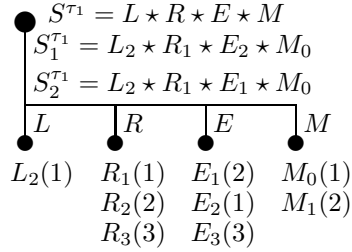
\put(00,00){\makebox(0,0)[bl] {Fig. 10. Structure of team
(\(\tau_{1}\))}}

\put(1,13){\makebox(0,8)[bl]{\(L_{2}(1)\)}}

\put(13,13){\makebox(0,8)[bl]{\(R_{1}(1)\)}}
\put(13,09){\makebox(0,8)[bl]{\(R_{2}(2)\)}}
\put(13,05){\makebox(0,8)[bl]{\(R_{3}(3)\)}}

\put(25,13){\makebox(0,8)[bl]{\(E_{1}(2)\)}}
\put(25,09){\makebox(0,8)[bl]{\(E_{2}(1)\)}}
\put(25,05){\makebox(0,8)[bl]{\(E_{3}(3)\)}}

\put(37,13){\makebox(0,8)[bl]{\(M_{0}(1)\)}}
\put(37,09){\makebox(0,8)[bl]{\(M_{1}(2)\)}}


\put(03,19){\circle*{2}} \put(15,19){\circle*{2}}
\put(27,19){\circle*{2}} \put(39,19){\circle*{2}}

\put(03,24){\line(0,-1){04}} \put(15,24){\line(0,-1){04}}
\put(27,24){\line(0,-1){04}} \put(39,24){\line(0,-1){04}}


\put(03,24){\line(1,0){36}}

\put(04,20.5){\makebox(0,8)[bl]{\(L\) }}
\put(16,20.5){\makebox(0,8)[bl]{\(R\) }}
\put(28,20.5){\makebox(0,8)[bl]{\(E\) }}
\put(40,20.5){\makebox(0,8)[bl]{\(M\) }}


\put(03,24){\line(0,1){10}} \put(03,35){\circle*{2.7}}

\put(06,35){\makebox(0,8)[bl] {\(S^{\tau_{1}} =
 L \star R \star E \star M \)}}

\put(04,30){\makebox(0,8)[bl] {\(S^{\tau_{1}}_{1} =
 L_{2} \star R_{1} \star E_{2} \star M_{0}  \)}}

\put(04,25){\makebox(0,8)[bl] {\(S^{\tau_{1}}_{2} =
 L_{2} \star R_{1} \star E_{1} \star M_{0}  \)}}

\end{picture}
%
\begin{picture}(54,42)

\put(06,37.5){\makebox(0,0)[bl]{Table 3. Compatibility
(\(\tau_{1}\))}}

\put(00,0){\line(1,0){54}} \put(00,30){\line(1,0){54}}
\put(00,36){\line(1,0){54}}

\put(00,0){\line(0,1){36}} \put(06,0){\line(0,1){36}}
\put(54,0){\line(0,1){36}}

\put(01,26){\makebox(0,0)[bl]{\(L_{2}\)}}

\put(01,22){\makebox(0,0)[bl]{\(R_{1}\)}}
\put(01,18){\makebox(0,0)[bl]{\(R_{2}\)}}
\put(01,14){\makebox(0,0)[bl]{\(R_{3}\)}}

\put(01,10){\makebox(0,0)[bl]{\(E_{1}\)}}
\put(01,06){\makebox(0,0)[bl]{\(E_{2}\)}}
\put(01,02){\makebox(0,0)[bl]{\(E_{3}\)}}

\put(12,30){\line(0,1){6}} \put(18,30){\line(0,1){6}}
\put(24,30){\line(0,1){6}} \put(30,30){\line(0,1){6}}
\put(36,30){\line(0,1){6}} \put(42,30){\line(0,1){6}}
\put(48,30){\line(0,1){6}}

\put(07,32){\makebox(0,0)[bl]{\(R_{1}\)}}
\put(13,32){\makebox(0,0)[bl]{\(R_{2}\)}}
\put(19,32){\makebox(0,0)[bl]{\(R_{3}\)}}
\put(25,32){\makebox(0,0)[bl]{\(E_{1}\)}}
\put(31,32){\makebox(0,0)[bl]{\(E_{2}\)}}
\put(37,32){\makebox(0,0)[bl]{\(E_{3}\)}}
\put(42.6,32){\makebox(0,0)[bl]{\(M_{0}\)}}
\put(48.6,32){\makebox(0,0)[bl]{\(M_{1}\)}}


\put(08,26){\makebox(0,0)[bl]{\(3\)}}
\put(14,26){\makebox(0,0)[bl]{\(2\)}}
\put(20,26){\makebox(0,0)[bl]{\(2\)}}
\put(26,26){\makebox(0,0)[bl]{\(3\)}}
\put(32,26){\makebox(0,0)[bl]{\(3\)}}
\put(38,26){\makebox(0,0)[bl]{\(1\)}}
\put(44,26){\makebox(0,0)[bl]{\(3\)}}
\put(50,26){\makebox(0,0)[bl]{\(1\)}}

\put(26,22){\makebox(0,0)[bl]{\(3\)}}
\put(32,22){\makebox(0,0)[bl]{\(2\)}}
\put(38,22){\makebox(0,0)[bl]{\(2\)}}
\put(44,22){\makebox(0,0)[bl]{\(3\)}}
\put(50,22){\makebox(0,0)[bl]{\(1\)}}

\put(26,18){\makebox(0,0)[bl]{\(2\)}}
\put(32,18){\makebox(0,0)[bl]{\(2\)}}
\put(38,18){\makebox(0,0)[bl]{\(1\)}}
\put(44,18){\makebox(0,0)[bl]{\(3\)}}
\put(50,18){\makebox(0,0)[bl]{\(3\)}}

\put(26,14){\makebox(0,0)[bl]{\(1\)}}
\put(32,14){\makebox(0,0)[bl]{\(2\)}}
\put(38,14){\makebox(0,0)[bl]{\(3\)}}
\put(44,14){\makebox(0,0)[bl]{\(2\)}}
\put(50,14){\makebox(0,0)[bl]{\(2\)}}

\put(44,10){\makebox(0,0)[bl]{\(3\)}}
\put(50,10){\makebox(0,0)[bl]{\(3\)}}

\put(44,06){\makebox(0,0)[bl]{\(3\)}}
\put(50,06){\makebox(0,0)[bl]{\(2\)}}

\put(44,02){\makebox(0,0)[bl]{\(3\)}}
\put(50,02){\makebox(0,0)[bl]{\(2\)}}

\end{picture}
\end{center}

 The resultant composite Pareto-efficient DAs are the following (Fig. 13):

 (a)
 \(S^{\tau_{1}}_{1} = L_{2} \star R_{1} \star E_{2} \star M_{0} \),
 \(N(S^{\tau_{1}}_{1})= (2;4,0,0)\);

  (b)
 \(S^{\tau_{1}}_{2} = L_{2} \star R_{1} \star E_{1} \star M_{0} \),
 \(N(S^{\tau_{1}}_{2})= (3;3,1,0)\).

\begin{center}
\begin{picture}(55,35)

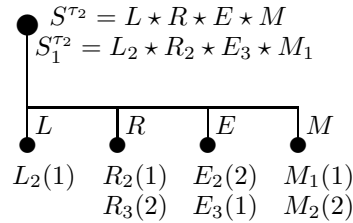
\put(00,00){\makebox(0,0)[bl] {Fig. 11. Structure of team
(\(\tau_{2}\))}}

\put(1,09){\makebox(0,8)[bl]{\(L_{2}(1)\)}}

\put(13,09){\makebox(0,8)[bl]{\(R_{2}(1)\)}}
\put(13,05){\makebox(0,8)[bl]{\(R_{3}(2)\)}}

\put(25,09){\makebox(0,8)[bl]{\(E_{2}(2)\)}}
\put(25,05){\makebox(0,8)[bl]{\(E_{3}(1)\)}}

\put(37,09){\makebox(0,8)[bl]{\(M_{1}(1)\)}}
\put(37,05){\makebox(0,8)[bl]{\(M_{2}(2)\)}}


\put(03,15){\circle*{2}} \put(15,15){\circle*{2}}
\put(27,15){\circle*{2}} \put(39,15){\circle*{2}}

\put(03,20){\line(0,-1){04}} \put(15,20){\line(0,-1){04}}
\put(27,20){\line(0,-1){04}} \put(39,20){\line(0,-1){04}}


\put(03,20){\line(1,0){36}}

\put(04,16.5){\makebox(0,8)[bl]{\(L\) }}
\put(16,16.5){\makebox(0,8)[bl]{\(R\) }}
\put(28,16.5){\makebox(0,8)[bl]{\(E\) }}
\put(40,16.5){\makebox(0,8)[bl]{\(M\) }}


\put(03,20){\line(0,1){10}} \put(03,31){\circle*{2.7}}

\put(06,31){\makebox(0,8)[bl] {\(S^{\tau_{2}} =
 L \star R \star E \star M \)}}

\put(04,26){\makebox(0,8)[bl] {\(S^{\tau_{2}}_{1} =
 L_{2} \star R_{2} \star E_{3} \star M_{1}
  \)}}


\end{picture}
%
\begin{picture}(42,34)

\put(00,29.5){\makebox(0,0)[bl]{Table 4. Compatibility
(\(\tau_{2}\))}}

\put(00,0){\line(1,0){42}} \put(00,22){\line(1,0){42}}
\put(00,28){\line(1,0){42}}

\put(00,0){\line(0,1){28}} \put(06,0){\line(0,1){28}}
\put(42,0){\line(0,1){28}}

\put(01,18){\makebox(0,0)[bl]{\(L_{2}\)}}

\put(01,14){\makebox(0,0)[bl]{\(R_{2}\)}}
\put(01,10){\makebox(0,0)[bl]{\(R_{3}\)}}

\put(01,06){\makebox(0,0)[bl]{\(E_{2}\)}}
\put(01,02){\makebox(0,0)[bl]{\(E_{3}\)}}

\put(12,22){\line(0,1){6}} \put(18,22){\line(0,1){6}}
\put(24,22){\line(0,1){6}} \put(30,22){\line(0,1){6}}
\put(36,22){\line(0,1){6}}

\put(07,24){\makebox(0,0)[bl]{\(R_{2}\)}}
\put(13,24){\makebox(0,0)[bl]{\(R_{3}\)}}
\put(19,24){\makebox(0,0)[bl]{\(E_{2}\)}}
\put(25,24){\makebox(0,0)[bl]{\(E_{3}\)}}
\put(30.6,24){\makebox(0,0)[bl]{\(M_{1}\)}}
\put(36.6,24){\makebox(0,0)[bl]{\(M_{2}\)}}


\put(08,18){\makebox(0,0)[bl]{\(3\)}}
\put(14,18){\makebox(0,0)[bl]{\(2\)}}
\put(20,18){\makebox(0,0)[bl]{\(2\)}}
\put(26,18){\makebox(0,0)[bl]{\(3\)}}
\put(32,18){\makebox(0,0)[bl]{\(3\)}}
\put(38,18){\makebox(0,0)[bl]{\(2\)}}

\put(20,14){\makebox(0,0)[bl]{\(2\)}}
\put(26,14){\makebox(0,0)[bl]{\(3\)}}
\put(32,14){\makebox(0,0)[bl]{\(3\)}}
\put(38,14){\makebox(0,0)[bl]{\(2\)}}

\put(20,10){\makebox(0,0)[bl]{\(2\)}}
\put(26,10){\makebox(0,0)[bl]{\(3\)}}
\put(32,10){\makebox(0,0)[bl]{\(3\)}}
\put(38,10){\makebox(0,0)[bl]{\(2\)}}


\put(38,06){\makebox(0,0)[bl]{\(2\)}}

\put(38,02){\makebox(0,0)[bl]{\(2\)}}

\end{picture}
\end{center}

 The resultant composite Pareto-efficient DA is the following (Fig. 13):

 \(S^{\tau_{2}}_{1} = L_{2} \star R_{2} \star E_{3} \star M_{1} \),
 \(N(S^{\tau_{2}}_{1})= (3;4,0,0)\).

\begin{center}
\begin{picture}(55,39)

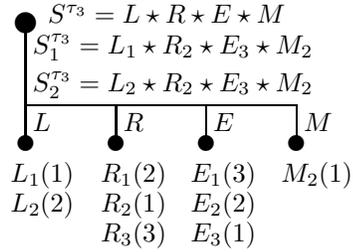
\put(00,00){\makebox(0,0)[bl] {Fig. 12. Structure of team
(\(\tau_{3}\))}}

\put(1,13){\makebox(0,8)[bl]{\(L_{1}(1)\)}}
\put(1,09){\makebox(0,8)[bl]{\(L_{2}(2)\)}}

\put(13,13){\makebox(0,8)[bl]{\(R_{1}(2)\)}}
\put(13,09){\makebox(0,8)[bl]{\(R_{2}(1)\)}}
\put(13,05){\makebox(0,8)[bl]{\(R_{3}(3)\)}}

\put(25,13){\makebox(0,8)[bl]{\(E_{1}(3)\)}}
\put(25,09){\makebox(0,8)[bl]{\(E_{2}(2)\)}}
\put(25,05){\makebox(0,8)[bl]{\(E_{3}(1)\)}}

\put(37,13){\makebox(0,8)[bl]{\(M_{2}(1)\)}}


\put(03,19){\circle*{2}} \put(15,19){\circle*{2}}
\put(27,19){\circle*{2}} \put(39,19){\circle*{2}}

\put(03,24){\line(0,-1){04}} \put(15,24){\line(0,-1){04}}
\put(27,24){\line(0,-1){04}} \put(39,24){\line(0,-1){04}}


\put(03,24){\line(1,0){36}}

\put(04,20.5){\makebox(0,8)[bl]{\(L\) }}
\put(16,20.5){\makebox(0,8)[bl]{\(R\) }}
\put(28,20.5){\makebox(0,8)[bl]{\(E\) }}
\put(40,20.5){\makebox(0,8)[bl]{\(M\) }}


\put(03,24){\line(0,1){10}} \put(03,35){\circle*{2.7}}

\put(06,35){\makebox(0,8)[bl] {\(S^{\tau_{3}} =
 L \star R \star E \star M \)}}

\put(04,30){\makebox(0,8)[bl] {\(S^{\tau_{3}}_{1} =
 L_{1} \star R_{2} \star E_{3} \star M_{2}\)}}

\put(04,25){\makebox(0,8)[bl] {\(S^{\tau_{3}}_{2} =
 L_{2} \star R_{2} \star E_{3} \star M_{2} \)}}

\end{picture}
%
\begin{picture}(48,46)

\put(02,41.5){\makebox(0,0)[bl]{Table 5. Compatibility
(\(\tau_{3}\))}}

\put(00,0){\line(1,0){48}} \put(00,34){\line(1,0){48}}
\put(00,40){\line(1,0){48}}

\put(00,0){\line(0,1){40}} \put(06,0){\line(0,1){40}}
\put(48,0){\line(0,1){40}}

\put(01,30){\makebox(0,0)[bl]{\(L_{1}\)}}
\put(01,26){\makebox(0,0)[bl]{\(L_{2}\)}}

\put(01,22){\makebox(0,0)[bl]{\(R_{1}\)}}
\put(01,18){\makebox(0,0)[bl]{\(R_{2}\)}}
\put(01,14){\makebox(0,0)[bl]{\(R_{3}\)}}

\put(01,10){\makebox(0,0)[bl]{\(E_{1}\)}}
\put(01,06){\makebox(0,0)[bl]{\(E_{2}\)}}
\put(01,02){\makebox(0,0)[bl]{\(E_{3}\)}}

\put(12,34){\line(0,1){6}} \put(18,34){\line(0,1){6}}
\put(24,34){\line(0,1){6}} \put(30,34){\line(0,1){6}}
\put(36,34){\line(0,1){6}} \put(42,34){\line(0,1){6}}

\put(07,36){\makebox(0,0)[bl]{\(R_{1}\)}}
\put(13,36){\makebox(0,0)[bl]{\(R_{2}\)}}
\put(19,36){\makebox(0,0)[bl]{\(R_{3}\)}}
\put(25,36){\makebox(0,0)[bl]{\(E_{1}\)}}
\put(31,36){\makebox(0,0)[bl]{\(E_{2}\)}}
\put(37,36){\makebox(0,0)[bl]{\(E_{3}\)}}
\put(42.6,36){\makebox(0,0)[bl]{\(M_{2}\)}}


\put(08,30){\makebox(0,0)[bl]{\(2\)}}
\put(14,30){\makebox(0,0)[bl]{\(2\)}}
\put(20,30){\makebox(0,0)[bl]{\(2\)}}
\put(26,30){\makebox(0,0)[bl]{\(2\)}}
\put(32,30){\makebox(0,0)[bl]{\(3\)}}
\put(38,30){\makebox(0,0)[bl]{\(3\)}}
\put(44,30){\makebox(0,0)[bl]{\(3\)}}

\put(08,26){\makebox(0,0)[bl]{\(2\)}}
\put(14,26){\makebox(0,0)[bl]{\(3\)}}
\put(20,26){\makebox(0,0)[bl]{\(2\)}}
\put(26,26){\makebox(0,0)[bl]{\(2\)}}
\put(32,26){\makebox(0,0)[bl]{\(3\)}}
\put(38,26){\makebox(0,0)[bl]{\(3\)}}
\put(44,26){\makebox(0,0)[bl]{\(3\)}}

\put(26,22){\makebox(0,0)[bl]{\(3\)}}
\put(32,22){\makebox(0,0)[bl]{\(2\)}}
\put(38,22){\makebox(0,0)[bl]{\(3\)}}
\put(44,22){\makebox(0,0)[bl]{\(2\)}}

\put(26,18){\makebox(0,0)[bl]{\(3\)}}
\put(32,18){\makebox(0,0)[bl]{\(3\)}}
\put(38,18){\makebox(0,0)[bl]{\(3\)}}
\put(44,18){\makebox(0,0)[bl]{\(3\)}}

\put(26,14){\makebox(0,0)[bl]{\(3\)}}
\put(32,14){\makebox(0,0)[bl]{\(3\)}}
\put(38,14){\makebox(0,0)[bl]{\(2\)}}
\put(44,14){\makebox(0,0)[bl]{\(3\)}}

\put(44,10){\makebox(0,0)[bl]{\(2\)}}

\put(44,06){\makebox(0,0)[bl]{\(3\)}}

\put(44,02){\makebox(0,0)[bl]{\(3\)}}

\end{picture}
\end{center}

 The resultant composite Pareto-efficient DAs are the following (Fig. 13):

 (a)
 \(S^{\tau_{3}}_{1} = L_{1} \star R_{2} \star E_{3} \star M_{2} \),
 \(N(S^{\tau_{3}}_{1})= (2;4,0,0)\);

 (b)
 \(S^{\tau_{3}}_{2} = L_{2} \star R_{2} \star E_{3} \star M_{2} \),
 \(N(S^{\tau_{3}}_{2})= (3;3,1,0)\).

 Table 6 contains compatibility estimates for the obtained local
 solutions.
%
 It is assumed local solutions have the same priorities (i.e., \(1\)).

 Thus, the final four-stage Pareto-efficient trajectory is (Fig. 14):

 \(\alpha^{team} = <
 S^{\tau_{0}}_{1} \star S^{\tau_{1}}_{2}  \star S^{\tau_{2}}_{1} \star S^{\tau_{3}}_{1} >\),
   \(N(\alpha^{team}) = (3;4,0,0)\).


\begin{center}
\begin{picture}(64,62)

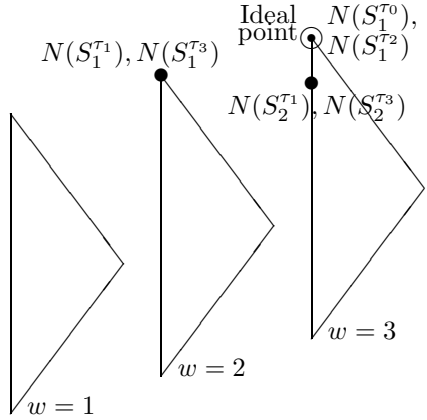
\put(001,00){\makebox(0,0)[bl]{Fig. 13. Poset of system quality}}

\put(00,06){\line(0,1){40}} \put(00,06){\line(3,4){15}}
\put(00,046){\line(3,-4){15}}

\put(20,011){\line(0,1){40}} \put(20,011){\line(3,4){15}}
\put(20,051){\line(3,-4){15}}

\put(40,016){\line(0,1){40}} \put(40,016){\line(3,4){15}}
\put(40,056){\line(3,-4){15}}



\put(40,50){\circle*{1.6}}
\put(28.8,45){\makebox(0,0)[bl]{\(N(S^{\tau_{1}}_{2}),
 N(S^{\tau_{3}}_{2})\)}}


\put(20,51){\circle*{1.6}}
\put(04,52){\makebox(0,0)[bl]{\(N(S^{\tau_{1}}_{1}),
 N(S^{\tau_{3}}_{1})\)}}


\put(40,56){\circle*{1}} \put(40,56){\circle{2.7}}

\put(30.5,58){\makebox(0,0)[bl]{Ideal}}
\put(30.2,55){\makebox(0,0)[bl]{point}}

\put(42,57){\makebox(0,0)[bl]{\(N(S^{\tau_{0}}_{1})\),}}
\put(42,53){\makebox(0,0)[bl]{\(N(S^{\tau_{2}}_{1})\)}}


\put(02.2,06){\makebox(0,0)[bl]{\(w=1\)}}
\put(22.2,11){\makebox(0,0)[bl]{\(w=2\)}}
\put(42.2,16){\makebox(0,0)[bl]{\(w=3\)}}

\end{picture}
%
\begin{picture}(42,34)

\put(02.4,30){\makebox(0,0)[bl]{Table 6. Compatibility}}

\put(00,0){\line(1,0){42}} \put(00,21){\line(1,0){42}}
\put(00,28){\line(1,0){42}}

\put(00,0){\line(0,1){28}} \put(07,0){\line(0,1){28}}
\put(42,0){\line(0,1){28}}

\put(01,16){\makebox(0,0)[bl]{\(S^{\tau_{0}}_{1}\)}}

\put(01,11){\makebox(0,0)[bl]{\(S^{\tau_{1}}_{1}\)}}
\put(01,06){\makebox(0,0)[bl]{\(S^{\tau_{1}}_{2}\)}}

\put(01,01){\makebox(0,0)[bl]{\(S^{\tau_{2}}_{1}\)}}

\put(14,21){\line(0,1){7}} \put(21,21){\line(0,1){7}}
\put(28,21){\line(0,1){7}} \put(35,21){\line(0,1){7}}

\put(08,22.5){\makebox(0,0)[bl]{\(S^{\tau_{1}}_{1}\)}}
\put(15,22.5){\makebox(0,0)[bl]{\(S^{\tau_{1}}_{2}\)}}
\put(22,22.5){\makebox(0,0)[bl]{\(S^{\tau_{2}}_{1}\)}}
\put(29,22.5){\makebox(0,0)[bl]{\(S^{\tau_{3}}_{1}\)}}
\put(36,22.5){\makebox(0,0)[bl]{\(S^{\tau_{3}}_{2}\)}}


\put(09,16.5){\makebox(0,0)[bl]{\(2\)}}
\put(16,16.5){\makebox(0,0)[bl]{\(3\)}}
\put(23,16.5){\makebox(0,0)[bl]{\(3\)}}
\put(30,16.5){\makebox(0,0)[bl]{\(3\)}}
\put(37,16.5){\makebox(0,0)[bl]{\(3\)}}

\put(23,11.5){\makebox(0,0)[bl]{\(3\)}}
\put(30,11.5){\makebox(0,0)[bl]{\(3\)}}
\put(37,11.5){\makebox(0,0)[bl]{\(3\)}}

\put(23,6.5){\makebox(0,0)[bl]{\(2\)}}
\put(30,6.5){\makebox(0,0)[bl]{\(3\)}}
\put(37,6.5){\makebox(0,0)[bl]{\(3\)}}

\put(30,1.5){\makebox(0,0)[bl]{\(3\)}}
\put(37,1.5){\makebox(0,0)[bl]{\(2\)}}

\end{picture}
\end{center}

\begin{center}
\begin{picture}(96,24)
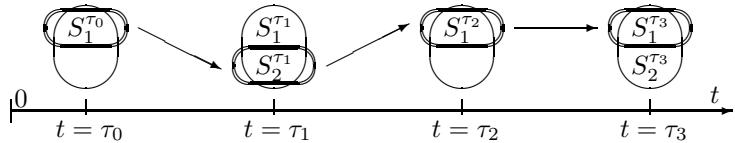
\put(25,00){\makebox(0,0)[bl] {Fig. 14. Trajectory of
 team}}

\put(0.5,9.5){\makebox(0,8)[bl]{\(0\)}}
\put(00,7.5){\line(0,1){3}}

\put(00,9){\vector(1,0){96}} \put(93,10){\makebox(0,8)[bl]{\(t\)}}

\put(6,5){\makebox(0,8)[bl]{\(t=\tau_{0}\)}}
\put(31,5){\makebox(0,8)[bl]{\(t=\tau_{1}\)}}
\put(56,5){\makebox(0,8)[bl]{\(t=\tau_{2}\)}}
\put(81,5){\makebox(0,8)[bl]{\(t=\tau_{3}\)}}

\put(10,8.5){\line(0,1){2}} \put(35,8.5){\line(0,1){2}}
\put(60,8.5){\line(0,1){2}} \put(85,8.5){\line(0,1){2}}


\put(10,20){\oval(11,5)} \put(10,20){\oval(10.5,4.5)}

\put(7.5,18){\makebox(0,8)[bl]{\(S^{\tau_{0}}_{1}\)}}

\put(10,17.5){\oval(08.5,11)}

 \put(17,20){\vector(2,-1){11}}


\put(35,15){\oval(11,5)} \put(35,15){\oval(10.5,4.5)}

\put(32.5,18){\makebox(0,8)[bl]{\(S^{\tau_{1}}_{1}\)}}
\put(32.5,13){\makebox(0,8)[bl]{\(S^{\tau_{1}}_{2}\)}}

\put(35,17.5){\oval(08.5,11)}


\put(42,15){\vector(2,1){11}}


\put(60,20){\oval(11,5)} \put(60,20){\oval(10.5,4.5)}

\put(57.5,18){\makebox(0,8)[bl]{\(S^{\tau_{2}}_{1}\)}}

\put(60,17.5){\oval(08.5,11)} \put(67,20){\vector(1,0){11}}



\put(85,20){\oval(11,5)} \put(85,20){\oval(10.5,4.5)}

\put(82.5,18){\makebox(0,8)[bl]{\(S^{\tau_{3}}_{1}\)}}
\put(82.5,13){\makebox(0,8)[bl]{\(S^{\tau_{3}}_{2}\)}}

\put(85,17.5){\oval(08.5,11)}


\end{picture}
\end{center}


\subsection{Tree-like Trajectory for Medical Treatment}

 Here, multi-stage design for medical treatment is examined.
 The example is based on the following:

 (i) basic tree-like structure for medical treatment
 for children asthma from \cite{levsok04}
 (a simplified version);

 (ii) top-level network as a decision tree (Fig. 6);

 (iii) modified tree-like structure
 of medical treatment for each node of the decision tree.

 The considered tree-like trajectory for medical treatment with
 'analysis/decision' nodes is depicted in Fig. 15.
  Each node of the trajectory is based on a simplified hierarchical
 structure of medical treatment for children asthma
 that has been suggested in
 \cite{levsok04}.

 Thus, the examined structure of the basic
 composite medical plan is the following
 (priorities of DAs are shown in  parentheses)
 (Fig. 16):

\begin{center}
\begin{picture}(76,43)

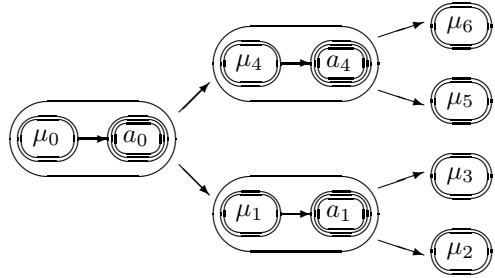
\put(01,00){\makebox(0,0)[bl]{Fig. 15. Decision-tree
 for medical treatment}}


\put(11,24){\oval(22,10)}

\put(05,24){\oval(8,6)} \put(05,24){\oval(7,5)}
\put(03,23){\makebox(0,8)[bl]{\(\mu_{0}\)}}

\put(09,24){\vector(1,0){4}}


\put(17,24){\oval(8,6)} \put(17,24){\oval(7,5)}
\put(17,24){\oval(6,4)}

\put(15,23){\makebox(0,8)[bl]{\(a_{0}\)}}

\put(22.5,27.5){\vector(1,1){4}} \put(22.5,20.5){\vector(1,-1){4}}


\put(38,14){\oval(22,10)}

\put(32,14){\oval(8,6)} \put(32,14){\oval(7,5)}
\put(30,13){\makebox(0,8)[bl]{\(\mu_{1}\)}}

\put(36,14){\vector(1,0){4}}


\put(44,14){\oval(8,6)} \put(44,14){\oval(7,5)}
\put(44,14){\oval(6,4)}

\put(42,13){\makebox(0,8)[bl]{\(a_{1}\)}}

\put(49,17.5){\vector(3,1){6}} \put(49,10.5){\vector(3,-1){6}}


\put(60,09){\oval(8,6)} \put(60,09){\oval(7,5)}
\put(58,08){\makebox(0,8)[bl]{\(\mu_{2}\)}}


\put(60,19){\oval(8,6)} \put(60,19){\oval(7,5)}
\put(58,18){\makebox(0,8)[bl]{\(\mu_{3}\)}}



\put(38,34){\oval(22,10)}

\put(32,34){\oval(8,6)} \put(32,34){\oval(7,5)}
\put(30,33){\makebox(0,8)[bl]{\(\mu_{4}\)}}

\put(36,34){\vector(1,0){4}}


\put(44,34){\oval(8,6)} \put(44,34){\oval(7,5)}
\put(44,34){\oval(6,4)}

\put(42,33){\makebox(0,8)[bl]{\(a_{4}\)}}

\put(49,37.5){\vector(3,1){6}} \put(49,30.5){\vector(3,-1){6}}



\put(60,29){\oval(8,6)} \put(60,29){\oval(7,5)}
\put(58,28){\makebox(0,8)[bl]{\(\mu_{5}\)}}


\put(60,39){\oval(8,6)} \put(60,39){\oval(7,5)}
\put(58,38){\makebox(0,8)[bl]{\(\mu_{6}\)}}




\end{picture}
\end{center}

\begin{center}
\begin{picture}(110,79)

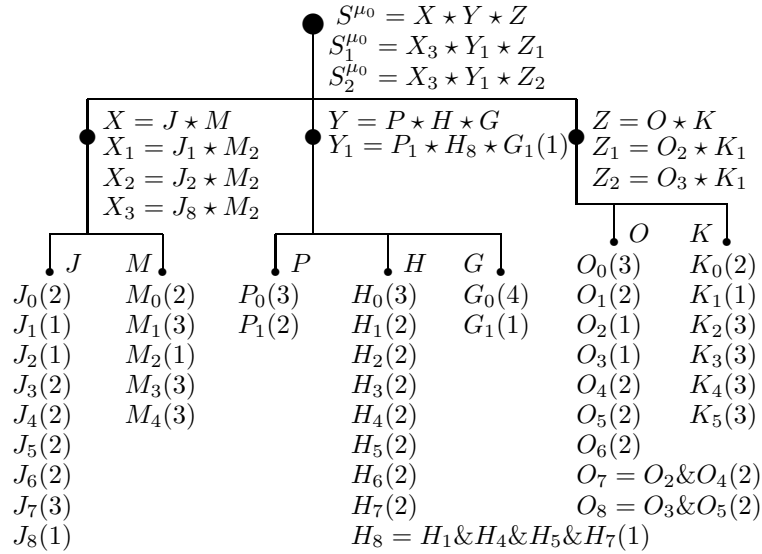
\put(02,00){\makebox(0,0)[bl] {Fig. 16. Hierarchical model of
 medical treatment plan \cite{lev06,levsok04} }}

\put(40,64){\line(0,1){11}} \put(40,75){\circle*{2.8}}

\put(43,75){\makebox(0,0)[bl]{\(S^{\mu_{0}}=X\star Y\star Z\)}}

\put(42,70){\makebox(0,0)[bl]{\(S_{1}^{\mu_{0}}=X_{3}\star
 Y_{1}\star Z_{1}\)}}

\put(42,66){\makebox(0,0)[bl]{\(S_{2}^{\mu_{0}}=X_{3}\star
 Y_{1}\star Z_{2}\)}}








\put(10,65){\line(1,0){65}}

\put(10,47){\line(0,1){18}} \put(10,60){\circle*{2}}

\put(12,61){\makebox(0,0)[bl]{\(X=J\star M\)}}

\put(12,57){\makebox(0,0)[bl]{\(X_{1}=J_{1}\star M_{2}\)}}
\put(12,53){\makebox(0,0)[bl]{\(X_{2}=J_{2}\star M_{2}\)}}
\put(12,49){\makebox(0,0)[bl]{\(X_{3}=J_{8}\star M_{2}\)}}




\put(40,47){\line(0,1){18}} \put(40,60){\circle*{2}}

\put(42,61){\makebox(0,0)[bl]{\(Y=P\star H\star G\)}}

\put(42,57){\makebox(0,0)[bl]{\(Y_{1}=P_{1}\star H_{8}\star
G_{1}(1)\)}}




\put(75,51){\line(0,1){14}} \put(75,60){\circle*{2}}

\put(77,61){\makebox(0,0)[bl]{\(Z=O\star K\)}}

\put(77,57){\makebox(0,0)[bl]{\(Z_{1}=O_{2}\star K_{1}\)}}
\put(77,53){\makebox(0,0)[bl]{\(Z_{2}=O_{3}\star K_{1}\)}}


\put(05,47){\line(1,0){15}} \put(05,42){\line(0,1){5}}
\put(05,42){\circle*{1}}

\put(7,42){\makebox(0,0)[bl]{\(J\)}}

\put(0,37){\makebox(0,0)[bl]{\(J_{0}(2)\)}}
\put(0,33){\makebox(0,0)[bl]{\(J_{1}(1)\)}}
\put(0,29){\makebox(0,0)[bl]{\(J_{2}(1)\)}}
\put(0,25){\makebox(0,0)[bl]{\(J_{3}(2)\)}}
\put(0,21){\makebox(0,0)[bl]{\(J_{4}(2)\)}}
\put(0,17){\makebox(0,0)[bl]{\(J_{5}(2)\)}}
\put(0,13){\makebox(0,0)[bl]{\(J_{6}(2)\)}}
\put(0,09){\makebox(0,0)[bl]{\(J_{7}(3)\)}}
\put(0,05){\makebox(0,0)[bl]{\(J_{8}(1)\)}}

\put(20,42){\line(0,1){5}} \put(20,42){\circle*{1}}

\put(15,42){\makebox(0,0)[bl]{\(M\)}}

\put(15,37){\makebox(0,0)[bl]{\(M_{0}(2)\)}}
\put(15,33){\makebox(0,0)[bl]{\(M_{1}(3)\)}}
\put(15,29){\makebox(0,0)[bl]{\(M_{2}(1)\)}}
\put(15,25){\makebox(0,0)[bl]{\(M_{3}(3)\)}}
\put(15,21){\makebox(0,0)[bl]{\(M_{4}(3)\)}}


\put(35,47){\line(1,0){30}} \put(35,42){\line(0,1){5}}
\put(35,42){\circle*{1}}

\put(37,42){\makebox(0,0)[bl]{\(P\)}}

\put(30,37){\makebox(0,0)[bl]{\(P_{0}(3)\)}}
\put(30,33){\makebox(0,0)[bl]{\(P_{1}(2)\)}}

\put(50,42){\line(0,1){5}} \put(50,42){\circle*{1}}

\put(52,42){\makebox(0,0)[bl]{\(H\)}}

\put(45,37){\makebox(0,0)[bl]{\(H_{0}(3)\)}}
\put(45,33){\makebox(0,0)[bl]{\(H_{1}(2)\)}}
\put(45,29){\makebox(0,0)[bl]{\(H_{2}(2)\)}}
\put(45,25){\makebox(0,0)[bl]{\(H_{3}(2)\)}}
\put(45,21){\makebox(0,0)[bl]{\(H_{4}(2)\)}}
\put(45,17){\makebox(0,0)[bl]{\(H_{5}(2)\)}}
\put(45,13){\makebox(0,0)[bl]{\(H_{6}(2)\)}}
\put(45,09){\makebox(0,0)[bl]{\(H_{7}(2)\)}}
\put(45,05){\makebox(0,0)[bl]{\(H_{8}=H_{1}\&H_{4}\&H_{5}\&H_{7}(1)\)}}

\put(65,42){\line(0,1){5}} \put(65,42){\circle*{1}}

\put(60,42){\makebox(0,0)[bl]{\(G\)}}

\put(60,37){\makebox(0,0)[bl]{\(G_{0}(4)\)}}
\put(60,33){\makebox(0,0)[bl]{\(G_{1}(1)\)}}
\put(75,51){\line(1,0){20}}

\put(80,46){\line(0,1){5}} \put(80,46){\circle*{1}}

\put(82,46){\makebox(0,0)[bl]{\(O\)}}

\put(75,41){\makebox(0,0)[bl]{\(O_{0}(3)\)}}
\put(75,37){\makebox(0,0)[bl]{\(O_{1}(2)\)}}
\put(75,33){\makebox(0,0)[bl]{\(O_{2}(1)\)}}
\put(75,29){\makebox(0,0)[bl]{\(O_{3}(1)\)}}
\put(75,25){\makebox(0,0)[bl]{\(O_{4}(2)\)}}
\put(75,21){\makebox(0,0)[bl]{\(O_{5}(2)\)}}
\put(75,17){\makebox(0,0)[bl]{\(O_{6}(2)\)}}
\put(75,13){\makebox(0,0)[bl]{\(O_{7}=O_{2}\&O_{4}(2)\)}}
\put(75,09){\makebox(0,0)[bl]{\(O_{8}=O_{3}\&O_{5}(2)\)}}

\put(95,46){\line(0,1){5}} \put(95,46){\circle*{1}}

\put(90,46){\makebox(0,0)[bl]{\(K\)}}

\put(90,41){\makebox(0,0)[bl]{\(K_{0}(2)\)}}
\put(90,37){\makebox(0,0)[bl]{\(K_{1}(1)\)}}
\put(90,33){\makebox(0,0)[bl]{\(K_{2}(3)\)}}
\put(90,29){\makebox(0,0)[bl]{\(K_{3}(3)\)}}
\put(90,25){\makebox(0,0)[bl]{\(K_{4}(3)\)}}
\put(90,21){\makebox(0,0)[bl]{\(K_{5}(3)\)}}

\end{picture}
\end{center}


 {\bf 0.} Medical plan ~\(S = X \star Y \star  Z \).

 {\bf 1.} Basic treatment  ~\(X = J \star M\):

 {\it 1.1.} Physical therapy  \(J \):~
  none \(J_{0}(2)\),
  massage \(J_{1}(2)\),
  inhalation \(J_{2}(2)\),
  sauna \(J_{3}(3)\),
  reflexological therapy \(J_{4}(3)\),
  laser-therapy \(J_{5}(3)\),
  massage for special centers/points \(J_{6}(3)\),
  reflexological therapy for special centers \(J_{7}(4)\),
  halo-cameras or salt mines \(J_{8}(2)\).

 {\it 1.2.} Drug treatment  \(M\):~
  none \(M_{0}(2)\),
  vitamins \(M_{1}(3)\),
  sodium chromoglycate (one month and two times in a year)  \(M_{2}(2)\),
  sodium chromoglycate (two months) \(M_{3}\)(3),
  sodium chromoglycate (three months) \(M_{4}(3)\).

 {\bf 2.} Psychological and ecological environment
 ~\(Y = P \star H \star G \):

 {\it 2.1.} Psychological climate \(P\):~
  none \(P_{0}(3)\),
  consulting of a psychologist \(P_{1}(2)\).

 {\it 2.2.} Home ecological environment \(H\):~
  none \(H_{0}(3)\),
  water cleaning \(H_{1}(1)\),
  to clean a book dust \(H_{2}(2)\),
  to take away cotton wool things (blanket, pillow, mattress) \(H_{3}(1)\),
  to take away carpets \(H_{4}(1)\),
  to exclude contacts with home animals \(H_{5}(2)\),
  to destroy cockroach environment \(H_{6}(2)\),
  to take away flowers \(H_{7}(1)\),
  aggregated alternative \(H_{8}=H_{1}\&H_{4}\&H_{5}\&H_{7}(1)\).

 {\it 2.2.} General ecological environment \(G\):~
 none \(G_{0}(4)\),
 improving the area of the residence \(G_{1}(1)\).

 {\bf 3.} Mode, rest and relaxation ~\(Z = O \star K \):

 {\it 3.1.} Mode \(O \):~
  none \(O_{0}(3)\),
  relaxation at the noon \(O_{1}(1)\),
  special physical actions (drainage, expectoration) \(O_{2}(1)\),
  sport (running, skiing, swimming) \(O_{3}(1)\),
  comfort shower-bath \(O_{4}(1)\),
  cold shower-bath \(O_{5}(2)\),
  the exclude electronic games \(O_{6}(2)\),
  aggregated alternative \(O_{7}=O_{2}\&O_{4}(2)\),
  aggregated alternative \(O_{8}=O_{3}\&O_{5}(2)\).

 {\it 3.2.} Relaxation/rest \(K \):~
 none \(K_{0}(2)\),
 a rest at forest-like environment \(K_{1}(2)\),
 a rest near see \(K_{2}(3)\),
 a rest at mountains \(K_{3}(4)\),
 special camps \(K_{4}(3)\),
 treatment in salt mines \(K_{5}(3)\).

~~

 In Fig. 16, the hierarchy (i.e., morphological structure)
 corresponds to logical point
 \(\mu_{0}\) (Fig. 15).
 Estimates of compatibility for DAs are presented in
 Table 7, Table 8, and Table 9 \cite{lev06,levsok04}
 (as simplified version, for all logical points
 \(\{\mu_{0},\mu_{1},\mu_{2},\mu_{3},\mu_{4},\mu_{5},\mu_{6}\}\)).
 Estimates of compatibility for DAs at the higher hierarchical level
 are presented in
 Table 10 (\(\mu_{0}\)).
 Table 11 contains descriptions of logical points
 including references to corresponding morphological structures
 (Fig. 17, Fig. 18, Fig. 19, and Fig. 20).

 For point \(\mu_{0}\) (Fig. 16),
 the resultant composite Pareto-efficient DAs are:

 (1) local Pareto-efficient solutions for subsystem \(X\):~
 \(X_{1} = J_{1}\star M_{2}\),  \(N(X_{1})= (3;2,0)\);
 \(X_{2} = J_{2}\star M_{2}\),  \(N(X_{2})= (3;2,0)\);
 \(X_{3} = J_{8}\star M_{2}\),  \(N(X_{3})= (3;2,0)\);

 (2) local Pareto-efficient solutions for subsystem \(Y\):~
 \(Y_{1} = P_{1}\star H_{8} \star G_{1}\),  \(N(Y_{1})= (3;2,1,0)\);

  (3) local Pareto-efficient solutions for subsystem \(Z\):~
 \(Z_{1} = O_{1}\star K_{2}\),  \(N(Z_{1})= (3;2,0,0)\);
 \(Z_{2} = O_{2}\star K_{2}\),  \(N(Z_{2})= (3;2,0,0)\).

 (4) final composite Pareto-efficient DAs for system  \(S\):~
 (a)
 \(S^{\mu_{0}}_{1} = X_{3} \star Y_{1} \star Z_{1}  \),
%
 (b)
 \(S^{\mu_{0}}_{2} = X_{3} \star Y_{1} \star Z_{2} \).

 For point \(\mu_{1}\) (Fig. 17),
 the resultant composite Pareto-efficient DA is:~
 \(S^{\mu_{1}}_{1}=P_{1}\star H_{8}\star G_{1}\),
 \(N(S^{\mu_{1}}_{1})= (3;3,0,0)\).

 For point \(\mu_{2}\) (Fig. 18, and for \(\mu_{5}\)),
 the resultant composite Pareto-efficient DAs are:~
 (a) \(S^{\mu_{2}}_{1}=O_{7}\star K_{1}\),
 \(N(S^{\mu_{2}}_{1})= (3;1,1,0)\);
 (b) \(S^{\mu_{2}}_{2}=O_{8}\star K_{1}\),
   \(N(S^{\mu_{2}}_{2})= (3;1,1,0)\).

 For point \(\mu_{3}\) (Fig. 19),
 the resultant composite Pareto-efficient DAs are:~
 (a) \(S^{\mu_{3}}_{1}=J_{1}\),
%
 (b) \(S^{\mu_{3}}_{2}=J_{2}\),

 (c) \(S^{\mu_{3}}_{3}=J_{8}\).

 For point \(\mu_{4}\) (Fig. 20, and for \(\mu_{6}\)),
 the resultant composite Pareto-efficient DAs are:~
 (a) \(S^{\mu_{4}}_{1}=J_{1}\star M_{1}\),
 \(N(S^{\mu_{2}}_{1})= (3;2,0)\);
 (b) \(S^{\mu_{4}}_{2}=J_{2}\star M_{1}\),
   \(N(S^{\mu_{2}}_{2})= (3;2,0)\);
 (c) \(S^{\mu_{4}}_{3}=J_{8}\star M_{1}\),
   \(N(S^{\mu_{2}}_{3})= (2;2,0)\).

 Table 12 contains descriptions of
 'analysis/decision' points.
 An example of the final tree solution is presented in Fig. 21.

\begin{center}
\begin{picture}(50,50)
\put(02,46){\makebox(0,0)[bl]{Table 7. Compatibility}}

\put(0,00){\line(1,0){41}} \put(0,38){\line(1,0){41}}
\put(0,44){\line(1,0){41}}

\put(0,00){\line(0,1){44}} \put(10,00){\line(0,1){44}}
\put(41,00){\line(0,1){44}}

\put(16,38){\line(0,1){6}} \put(22,38){\line(0,1){6}}
\put(28,38){\line(0,1){6}} \put(34,38){\line(0,1){6}}


\put(1,34){\makebox(0,0)[bl]{\(J_{0}\)}}
\put(1,30){\makebox(0,0)[bl]{\(J_{1}\)}}
\put(1,26){\makebox(0,0)[bl]{\(J_{2}\)}}
\put(1,22){\makebox(0,0)[bl]{\(J_{3}\)}}
\put(1,18){\makebox(0,0)[bl]{\(J_{4}\)}}
\put(1,14){\makebox(0,0)[bl]{\(J_{5}\)}}
\put(1,10){\makebox(0,0)[bl]{\(J_{6}\)}}
\put(1,06){\makebox(0,0)[bl]{\(J_{7}\)}}
\put(1,02){\makebox(0,0)[bl]{\(J_{8}\)}}


\put(11,40){\makebox(0,0)[bl]{\(M_{0}\)}}
\put(17,40){\makebox(0,0)[bl]{\(M_{1}\)}}
\put(23,40){\makebox(0,0)[bl]{\(M_{2}\)}}
\put(29,40){\makebox(0,0)[bl]{\(M_{3}\)}}
\put(35,40){\makebox(0,0)[bl]{\(M_{4}\)}}


\put(13,34){\makebox(0,0)[bl]{\(0\)}}
\put(19,34){\makebox(0,0)[bl]{\(3\)}}
\put(25,34){\makebox(0,0)[bl]{\(3\)}}
\put(31,34){\makebox(0,0)[bl]{\(3\)}}
\put(37,34){\makebox(0,0)[bl]{\(3\)}}

\put(13,30){\makebox(0,0)[bl]{\(3\)}}
\put(19,30){\makebox(0,0)[bl]{\(3\)}}
\put(25,30){\makebox(0,0)[bl]{\(3\)}}
\put(31,30){\makebox(0,0)[bl]{\(3\)}}
\put(37,30){\makebox(0,0)[bl]{\(3\)}}

\put(13,26){\makebox(0,0)[bl]{\(3\)}}
\put(19,26){\makebox(0,0)[bl]{\(3\)}}
\put(25,26){\makebox(0,0)[bl]{\(3\)}}
\put(31,26){\makebox(0,0)[bl]{\(3\)}}
\put(37,26){\makebox(0,0)[bl]{\(3\)}}

\put(13,22){\makebox(0,0)[bl]{\(3\)}}
\put(19,22){\makebox(0,0)[bl]{\(3\)}}
\put(25,22){\makebox(0,0)[bl]{\(3\)}}
\put(31,22){\makebox(0,0)[bl]{\(2\)}}
\put(37,22){\makebox(0,0)[bl]{\(2\)}}

\put(13,18){\makebox(0,0)[bl]{\(3\)}}
\put(19,18){\makebox(0,0)[bl]{\(3\)}}
\put(25,18){\makebox(0,0)[bl]{\(3\)}}
\put(31,18){\makebox(0,0)[bl]{\(2\)}}
\put(37,18){\makebox(0,0)[bl]{\(2\)}}

\put(13,14){\makebox(0,0)[bl]{\(3\)}}
\put(19,14){\makebox(0,0)[bl]{\(3\)}}
\put(25,14){\makebox(0,0)[bl]{\(3\)}}
\put(31,14){\makebox(0,0)[bl]{\(2\)}}
\put(37,14){\makebox(0,0)[bl]{\(2\)}}

\put(13,10){\makebox(0,0)[bl]{\(3\)}}
\put(19,10){\makebox(0,0)[bl]{\(3\)}}
\put(25,10){\makebox(0,0)[bl]{\(3\)}}
\put(31,10){\makebox(0,0)[bl]{\(3\)}}
\put(37,10){\makebox(0,0)[bl]{\(3\)}}

\put(13,06){\makebox(0,0)[bl]{\(3\)}}
\put(19,06){\makebox(0,0)[bl]{\(3\)}}
\put(25,06){\makebox(0,0)[bl]{\(3\)}}
\put(31,06){\makebox(0,0)[bl]{\(3\)}}
\put(37,06){\makebox(0,0)[bl]{\(3\)}}

\put(13,02){\makebox(0,0)[bl]{\(3\)}}
\put(19,02){\makebox(0,0)[bl]{\(2\)}}
\put(25,02){\makebox(0,0)[bl]{\(2\)}}
\put(31,02){\makebox(0,0)[bl]{\(2\)}}
\put(37,02){\makebox(0,0)[bl]{\(2\)}}

\end{picture}
\begin{picture}(47,50)
\put(04.5,46){\makebox(0,0)[bl]{Table 8. Compatibility}}

\put(0,00){\line(1,0){47}} \put(0,38){\line(1,0){47}}
\put(0,44){\line(1,0){47}}

\put(0,00){\line(0,1){44}} \put(10,00){\line(0,1){44}}
\put(47,00){\line(0,1){44}}

\put(16,38){\line(0,1){6}} \put(22,38){\line(0,1){6}}
\put(28,38){\line(0,1){6}} \put(34,38){\line(0,1){6}}
\put(40,38){\line(0,1){6}}


\put(1,34){\makebox(0,0)[bl]{\(O_{0}\)}}
\put(1,30){\makebox(0,0)[bl]{\(O_{1}\)}}
\put(1,26){\makebox(0,0)[bl]{\(O_{2}\)}}
\put(1,22){\makebox(0,0)[bl]{\(O_{3}\)}}
\put(1,18){\makebox(0,0)[bl]{\(O_{4}\)}}
\put(1,14){\makebox(0,0)[bl]{\(O_{5}\)}}
\put(1,10){\makebox(0,0)[bl]{\(O_{6}\)}}
\put(1,06){\makebox(0,0)[bl]{\(O_{7}\)}}
\put(1,02){\makebox(0,0)[bl]{\(O_{8}\)}}


\put(11,40){\makebox(0,0)[bl]{\(K_{0}\)}}
\put(17,40){\makebox(0,0)[bl]{\(K_{1}\)}}
\put(23,40){\makebox(0,0)[bl]{\(K_{2}\)}}
\put(29,40){\makebox(0,0)[bl]{\(K_{3}\)}}
\put(35,40){\makebox(0,0)[bl]{\(K_{4}\)}}
\put(41,40){\makebox(0,0)[bl]{\(K_{5}\)}}


\put(13,34){\makebox(0,0)[bl]{\(0\)}}
\put(19,34){\makebox(0,0)[bl]{\(3\)}}
\put(25,34){\makebox(0,0)[bl]{\(3\)}}
\put(31,34){\makebox(0,0)[bl]{\(3\)}}
\put(37,34){\makebox(0,0)[bl]{\(3\)}}
\put(43,34){\makebox(0,0)[bl]{\(3\)}}

\put(13,30){\makebox(0,0)[bl]{\(3\)}}
\put(19,30){\makebox(0,0)[bl]{\(3\)}}
\put(25,30){\makebox(0,0)[bl]{\(3\)}}
\put(31,30){\makebox(0,0)[bl]{\(3\)}}
\put(37,30){\makebox(0,0)[bl]{\(3\)}}
\put(43,30){\makebox(0,0)[bl]{\(3\)}}

\put(13,26){\makebox(0,0)[bl]{\(3\)}}
\put(19,26){\makebox(0,0)[bl]{\(3\)}}
\put(25,26){\makebox(0,0)[bl]{\(3\)}}
\put(31,26){\makebox(0,0)[bl]{\(3\)}}
\put(37,26){\makebox(0,0)[bl]{\(2\)}}
\put(43,26){\makebox(0,0)[bl]{\(3\)}}

\put(13,22){\makebox(0,0)[bl]{\(3\)}}
\put(19,22){\makebox(0,0)[bl]{\(3\)}}
\put(25,22){\makebox(0,0)[bl]{\(3\)}}
\put(31,22){\makebox(0,0)[bl]{\(3\)}}
\put(37,22){\makebox(0,0)[bl]{\(3\)}}
\put(43,22){\makebox(0,0)[bl]{\(3\)}}

\put(13,18){\makebox(0,0)[bl]{\(3\)}}
\put(19,18){\makebox(0,0)[bl]{\(2\)}}
\put(25,18){\makebox(0,0)[bl]{\(2\)}}
\put(31,18){\makebox(0,0)[bl]{\(2\)}}
\put(37,18){\makebox(0,0)[bl]{\(2\)}}
\put(43,18){\makebox(0,0)[bl]{\(2\)}}

\put(13,14){\makebox(0,0)[bl]{\(3\)}}
\put(19,14){\makebox(0,0)[bl]{\(3\)}}
\put(25,14){\makebox(0,0)[bl]{\(3\)}}
\put(31,14){\makebox(0,0)[bl]{\(3\)}}
\put(37,14){\makebox(0,0)[bl]{\(3\)}}
\put(43,14){\makebox(0,0)[bl]{\(3\)}}

\put(13,10){\makebox(0,0)[bl]{\(3\)}}
\put(19,10){\makebox(0,0)[bl]{\(3\)}}
\put(25,10){\makebox(0,0)[bl]{\(3\)}}
\put(31,10){\makebox(0,0)[bl]{\(3\)}}
\put(37,10){\makebox(0,0)[bl]{\(3\)}}
\put(43,10){\makebox(0,0)[bl]{\(3\)}}

\put(13,06){\makebox(0,0)[bl]{\(3\)}}
\put(19,06){\makebox(0,0)[bl]{\(3\)}}
\put(25,06){\makebox(0,0)[bl]{\(3\)}}
\put(31,06){\makebox(0,0)[bl]{\(3\)}}
\put(37,06){\makebox(0,0)[bl]{\(3\)}}
\put(43,06){\makebox(0,0)[bl]{\(3\)}}

\put(13,02){\makebox(0,0)[bl]{\(3\)}}
\put(19,02){\makebox(0,0)[bl]{\(3\)}}
\put(25,02){\makebox(0,0)[bl]{\(3\)}}
\put(31,02){\makebox(0,0)[bl]{\(3\)}}
\put(37,02){\makebox(0,0)[bl]{\(3\)}}
\put(43,02){\makebox(0,0)[bl]{\(3\)}}

\end{picture}
\end{center}

\begin{center}
\begin{picture}(78,30)

\put(16,26){\makebox(0,0)[bl]{Table 9. Compatibility}}

\put(00,00){\line(1,0){77}} \put(00,18){\line(1,0){77}}
\put(00,24){\line(1,0){77}}

\put(00,00){\line(0,1){24}} \put(10,00){\line(0,1){24}}
\put(77,00){\line(0,1){24}}

\put(16,18){\line(0,1){6}} \put(22,18){\line(0,1){6}}
\put(28,18){\line(0,1){6}} \put(34,18){\line(0,1){6}}
\put(40,18){\line(0,1){6}} \put(46,18){\line(0,1){6}}
\put(52,18){\line(0,1){6}} \put(58,18){\line(0,1){6}}
\put(64,18){\line(0,1){6}} \put(70,18){\line(0,1){6}}

\put(1,14){\makebox(0,0)[bl]{\(P_{0}\)}}
\put(1,10){\makebox(0,0)[bl]{\(P_{1}\)}}
\put(1,06){\makebox(0,0)[bl]{\(G_{0}\)}}
\put(1,02){\makebox(0,0)[bl]{\(G_{1}\)}}


\put(11,20){\makebox(0,0)[bl]{\(G_{0}\)}}
\put(17,20){\makebox(0,0)[bl]{\(G_{1}\)}}
\put(23,20){\makebox(0,0)[bl]{\(H_{0}\)}}
\put(29,20){\makebox(0,0)[bl]{\(H_{1}\)}}
\put(35,20){\makebox(0,0)[bl]{\(H_{2}\)}}
\put(41,20){\makebox(0,0)[bl]{\(H_{3}\)}}
\put(47,20){\makebox(0,0)[bl]{\(H_{4}\)}}
\put(53,20){\makebox(0,0)[bl]{\(H_{5}\)}}
\put(59,20){\makebox(0,0)[bl]{\(H_{6}\)}}
\put(65,20){\makebox(0,0)[bl]{\(H_{7}\)}}
\put(71,20){\makebox(0,0)[bl]{\(H_{8}\)}}


\put(13,14){\makebox(0,0)[bl]{\(0\)}}
\put(19,14){\makebox(0,0)[bl]{\(3\)}}
\put(25,14){\makebox(0,0)[bl]{\(0\)}}
\put(31,14){\makebox(0,0)[bl]{\(3\)}}
\put(37,14){\makebox(0,0)[bl]{\(3\)}}
\put(43,14){\makebox(0,0)[bl]{\(2\)}}
\put(49,14){\makebox(0,0)[bl]{\(3\)}}
\put(55,14){\makebox(0,0)[bl]{\(2\)}}
\put(61,14){\makebox(0,0)[bl]{\(3\)}}
\put(67,14){\makebox(0,0)[bl]{\(3\)}}
\put(73,14){\makebox(0,0)[bl]{\(3\)}}

\put(13,10){\makebox(0,0)[bl]{\(3\)}}
\put(19,10){\makebox(0,0)[bl]{\(3\)}}
\put(25,10){\makebox(0,0)[bl]{\(2\)}}
\put(31,10){\makebox(0,0)[bl]{\(3\)}}
\put(37,10){\makebox(0,0)[bl]{\(3\)}}
\put(43,10){\makebox(0,0)[bl]{\(2\)}}
\put(49,10){\makebox(0,0)[bl]{\(3\)}}
\put(55,10){\makebox(0,0)[bl]{\(2\)}}
\put(61,10){\makebox(0,0)[bl]{\(3\)}}
\put(67,10){\makebox(0,0)[bl]{\(3\)}}
\put(73,10){\makebox(0,0)[bl]{\(3\)}}

\put(25,06){\makebox(0,0)[bl]{\(0\)}}
\put(31,06){\makebox(0,0)[bl]{\(3\)}}
\put(37,06){\makebox(0,0)[bl]{\(3\)}}
\put(43,06){\makebox(0,0)[bl]{\(2\)}}
\put(49,06){\makebox(0,0)[bl]{\(3\)}}
\put(55,06){\makebox(0,0)[bl]{\(2\)}}
\put(61,06){\makebox(0,0)[bl]{\(3\)}}
\put(67,06){\makebox(0,0)[bl]{\(3\)}}
\put(73,06){\makebox(0,0)[bl]{\(3\)}}

\put(25,02){\makebox(0,0)[bl]{\(2\)}}
\put(31,02){\makebox(0,0)[bl]{\(3\)}}
\put(37,02){\makebox(0,0)[bl]{\(3\)}}
\put(43,02){\makebox(0,0)[bl]{\(2\)}}
\put(49,02){\makebox(0,0)[bl]{\(3\)}}
\put(55,02){\makebox(0,0)[bl]{\(2\)}}
\put(61,02){\makebox(0,0)[bl]{\(3\)}}
\put(67,02){\makebox(0,0)[bl]{\(3\)}}
\put(73,02){\makebox(0,0)[bl]{\(3\)}}

\end{picture}
\begin{picture}(37,30)
\put(00,26){\makebox(0,0)[bl] {Table 10. Compatibility}}

\put(00,00){\line(1,0){29}} \put(00,18){\line(1,0){29}}
\put(00,24){\line(1,0){29}}

\put(00,00){\line(0,1){24}} \put(10,00){\line(0,1){24}}
\put(29,00){\line(0,1){24}}

\put(16,18){\line(0,1){6}} \put(22,18){\line(0,1){6}}

\put(1,14){\makebox(0,0)[bl]{\(X_{1}\)}}
\put(1,10){\makebox(0,0)[bl]{\(X_{2}\)}}
\put(1,06){\makebox(0,0)[bl]{\(X_{3}\)}}
\put(1,02){\makebox(0,0)[bl]{\(Y_{1}\)}}

\put(11,20){\makebox(0,0)[bl]{\(Y_{1}\)}}
\put(17,20){\makebox(0,0)[bl]{\(Z_{1}\)}}
\put(23,20){\makebox(0,0)[bl]{\(Z_{2}\)}}

\put(13,14){\makebox(0,0)[bl]{\(3\)}}
\put(19,14){\makebox(0,0)[bl]{\(2\)}}
\put(25,14){\makebox(0,0)[bl]{\(2\)}}

\put(13,10){\makebox(0,0)[bl]{\(2\)}}
\put(19,10){\makebox(0,0)[bl]{\(2\)}}
\put(25,10){\makebox(0,0)[bl]{\(2\)}}

\put(13,06){\makebox(0,0)[bl]{\(3\)}}
\put(19,06){\makebox(0,0)[bl]{\(3\)}}
\put(25,06){\makebox(0,0)[bl]{\(3\)}}

\put(19,02){\makebox(0,0)[bl]{\(3\)}}
\put(25,02){\makebox(0,0)[bl]{\(3\)}}

\end{picture}
\end{center}

\begin{center}
\begin{picture}(60,42)

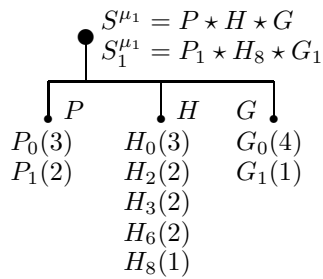
\put(00,00){\makebox(0,0)[bl] {Fig. 17. Treatment
 for point \(\mu_{1}\) }}


\put(10,31){\line(0,1){6}} \put(10,37){\circle*{2}}

\put(12,38){\makebox(0,0)[bl]{\(S^{\mu_{1}}=P\star H\star G\)}}

\put(12,33){\makebox(0,0)[bl]{\(S^{\mu_{1}}_{1}=P_{1}\star
 H_{8}\star G_{1}\)}}




\put(05,31){\line(1,0){30}}

\put(05,26){\line(0,1){5}} \put(05,26){\circle*{1}}

\put(07,26){\makebox(0,0)[bl]{\(P\)}}

\put(00,21){\makebox(0,0)[bl]{\(P_{0}(3)\)}}
\put(00,17){\makebox(0,0)[bl]{\(P_{1}(2)\)}}


\put(20,26){\line(0,1){5}} \put(20,26){\circle*{1}}

\put(22,26){\makebox(0,0)[bl]{\(H\)}}

\put(15,21){\makebox(0,0)[bl]{\(H_{0}(3)\)}}
\put(15,17){\makebox(0,0)[bl]{\(H_{2}(2)\)}}
\put(15,13){\makebox(0,0)[bl]{\(H_{3}(2)\)}}
\put(15,09){\makebox(0,0)[bl]{\(H_{6}(2)\)}}
\put(15,05){\makebox(0,0)[bl]{\(H_{8}(1)\)}}


\put(35,26){\line(0,1){5}} \put(35,26){\circle*{1}}

\put(30,26){\makebox(0,0)[bl]{\(G\)}}

\put(30,21){\makebox(0,0)[bl]{\(G_{0}(4)\)}}
\put(30,17){\makebox(0,0)[bl]{\(G_{1}(1)\)}}

\end{picture}
\begin{picture}(50,41)

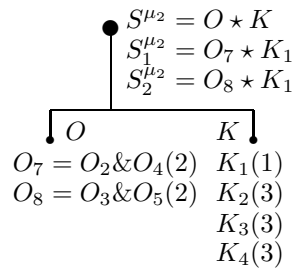
\put(00,00){\makebox(0,0)[bl] {Fig. 18. Treatment
  for
 point \(\mu_{2}\)}}


\put(13,26){\line(0,1){11}} \put(13,37){\circle*{2}}

\put(15,37){\makebox(0,0)[bl]{\(S^{\mu_{2}} =O\star K\)}}


\put(15,32){\makebox(0,0)[bl]{\(S^{\mu_{2}}_{1}=O_{7}\star
 K_{1}\)}}

\put(15,28){\makebox(0,0)[bl]{\(S^{\mu_{2}}_{2}=O_{8}\star
 K_{1}\)}}


\put(05,26){\line(1,0){27}}

\put(05,22){\line(0,1){4}} \put(05,22){\circle*{1}}

\put(7,22){\makebox(0,0)[bl]{\(O\)}}


\put(0,17){\makebox(0,0)[bl]{\(O_{7}=O_{2}\&O_{4}(2)\)}}
\put(0,13){\makebox(0,0)[bl]{\(O_{8}=O_{3}\&O_{5}(2)\)}}

\put(32,22){\line(0,1){4}} \put(32,22){\circle*{1}}

\put(27,22){\makebox(0,0)[bl]{\(K\)}}

\put(27,17){\makebox(0,0)[bl]{\(K_{1}(1)\)}}
\put(27,13){\makebox(0,0)[bl]{\(K_{2}(3)\)}}
\put(27,09){\makebox(0,0)[bl]{\(K_{3}(3)\)}}
\put(27,05){\makebox(0,0)[bl]{\(K_{4}(3)\)}}

\end{picture}
\end{center}

\begin{center}
\begin{picture}(60,39)

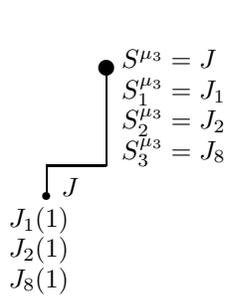
\put(00,00){\makebox(0,0)[bl] {Fig. 19. Treatment
  for
 point \(\mu_{3}\)}}


\put(13,22){\line(0,1){13}} \put(13,35){\circle*{2}}

\put(15,35){\makebox(0,0)[bl]{\(S^{\mu_{3}} =J\)}}

\put(15,30){\makebox(0,0)[bl]{\(S^{\mu_{3}}_{1}=J_{1}\)}}

\put(15,26){\makebox(0,0)[bl]{\(S^{\mu_{3}}_{2}=J_{2}\)}}

\put(15,22){\makebox(0,0)[bl]{\(S^{\mu_{3}}_{3}=J_{8}\)}}


\put(05,22){\line(1,0){08}}

\put(05,18){\line(0,1){4}} \put(05,18){\circle*{1}}

\put(7,18){\makebox(0,0)[bl]{\(J\)}}

\put(0,13){\makebox(0,0)[bl]{\(J_{1}(1)\)}}
\put(0,09){\makebox(0,0)[bl]{\(J_{2}(1)\)}}
\put(0,05){\makebox(0,0)[bl]{\(J_{8}(1)\)}}


\end{picture}
%
\begin{picture}(50,45)

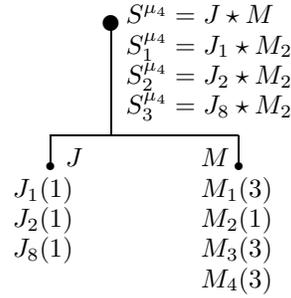
\put(00,00){\makebox(0,0)[bl] {Fig. 20. Treatment
  for
 point \(\mu_{4}\)}}


\put(13,26){\line(0,1){15}} \put(13,41){\circle*{2}}

\put(15,41){\makebox(0,0)[bl]{\(S^{\mu_{4}} =J\star M\)}}

\put(15,36){\makebox(0,0)[bl]{\(S^{\mu_{4}}_{1}=J_{1}\star
 M_{2}\)}}

\put(15,32){\makebox(0,0)[bl]{\(S^{\mu_{4}}_{2}=J_{2}\star
 M_{2}\)}}

\put(15,28){\makebox(0,0)[bl]{\(S^{\mu_{4}}_{3}=J_{8}\star
 M_{2}\)}}


\put(05,26){\line(1,0){25}}

\put(05,22){\line(0,1){4}} \put(05,22){\circle*{1}}

\put(7,22){\makebox(0,0)[bl]{\(J\)}}

\put(0,17){\makebox(0,0)[bl]{\(J_{1}(1)\)}}
\put(0,13){\makebox(0,0)[bl]{\(J_{2}(1)\)}}
\put(0,09){\makebox(0,0)[bl]{\(J_{8}(1)\)}}


\put(30,22){\line(0,1){4}} \put(30,22){\circle*{1}}

\put(25,22){\makebox(0,0)[bl]{\(M\)}}

\put(25,17){\makebox(0,0)[bl]{\(M_{1}(3)\)}}
\put(25,13){\makebox(0,0)[bl]{\(M_{2}(1)\)}}
\put(25,09){\makebox(0,0)[bl]{\(M_{3}(3)\)}}
\put(25,05){\makebox(0,0)[bl]{\(M_{4}(3)\)}}

\end{picture}
\end{center}

\begin{center}
\begin{picture}(60,82)

\put(09,78){\makebox(0,0)[bl] {Table 11. Logical points}}

\put(00,00){\line(1,0){58}} \put(0,66.5){\line(1,0){58}}
\put(00,76){\line(1,0){58}}

\put(00,00){\line(0,1){76}} \put(13,00){\line(0,1){76}}
\put(58,00){\line(0,1){76}}


\put(01,72){\makebox(0,0)[bl] {Logical}}
\put(01,68){\makebox(0,0)[bl] {point}}

\put(26,72){\makebox(0,0)[bl] {Description}}


\put(01,62){\makebox(0,0)[bl]{\(\mu_{0}\)}}
\put(14,61.5){\makebox(0,0)[bl]{Basic (fool) treatment}}
\put(14,58){\makebox(0,0)[bl]{(Fig. 16)}}


\put(01,54){\makebox(0,0)[bl]{\(\mu_{1}\)}}
\put(14,54){\makebox(0,0)[bl]{Treatment by environment}}
\put(14,50){\makebox(0,0)[bl]{(Fig. 17)}}


\put(01,46){\makebox(0,0)[bl]{\(\mu_{2}\)}}

\put(14,46){\makebox(0,0)[bl]{Additional treatment}}
\put(14,42){\makebox(0,0)[bl]{by relaxation (Fig. 18)}}


\put(01,38){\makebox(0,0)[bl]{\(\mu_{3}\)}}
\put(14,38){\makebox(0,0)[bl]{Additional physical therapy}}
\put(14,34){\makebox(0,0)[bl]{(Fig. 19)}}


\put(01,30){\makebox(0,0)[bl]{\(\mu_{4}\)}}

\put(14,30){\makebox(0,0)[bl]{Additional physical therapy}}
\put(14,26){\makebox(0,0)[bl]{and drug treatment}}
\put(14,22){\makebox(0,0)[bl]{(Fig. 20)}}


\put(01,18){\makebox(0,0)[bl]{\(\mu_{5}\)}}

\put(14,18){\makebox(0,0)[bl]{Additional treatment}}
\put(14,14){\makebox(0,0)[bl]{by environment (Fig. 18)}}


\put(01,10){\makebox(0,0)[bl]{\(\mu_{6}\)}}

\put(14,10){\makebox(0,0)[bl]{Additional physical therapy }}
\put(14,06){\makebox(0,0)[bl]{and drug treatment}}
\put(14,02){\makebox(0,0)[bl]{(Fig. 20)}}

\end{picture}
%
\begin{picture}(56,82)


\put(00,56){\makebox(0,0)[bl] {Table 12. 'Analysis/decision'
 points}}

\put(00,00){\line(1,0){56}} \put(00,40.8){\line(1,0){56}}
\put(00,54){\line(1,0){56}}

\put(00,00){\line(0,1){54}} \put(16.5,00){\line(0,1){54}}
\put(56,00){\line(0,1){54}}


\put(0.5,49){\makebox(0,0)[bl] {'Analysis/}}
\put(01,46){\makebox(0,0)[bl] {decision'}}
\put(01,42){\makebox(0,0)[bl] {point}}

\put(25,49){\makebox(0,0)[bl] {Description}}


\put(01,36){\makebox(0,0)[bl]{\(a_{0}\)}}

\put(17,36){\makebox(0,0)[bl]{(i) good results, go to
 \(\mu_{1}\)}}

\put(17,32){\makebox(0,0)[bl]{(ii) not sufficient results,}}
\put(22,28.5){\makebox(0,0)[bl]{go to \(\mu_{4}\)}}


\put(01,23){\makebox(0,0)[bl]{\(a_{1}\)}}

\put(17,23){\makebox(0,0)[bl]{(i) good results, go to
 \(\mu_{2}\)}}

\put(17,19){\makebox(0,0)[bl]{(ii) not sufficient results,}}
\put(22,15.5){\makebox(0,0)[bl]{go to \(\mu_{3}\)}}


\put(01,10){\makebox(0,0)[bl]{\(a_{4}\)}}

\put(17,10){\makebox(0,0)[bl]{(i) good results, go to
 \(\mu_{5}\)}}

\put(17,06){\makebox(0,0)[bl]{(ii) not sufficient results,}}
\put(22,02.5){\makebox(0,0)[bl]{go to \(\mu_{6}\)}}

\end{picture}
\end{center}


\begin{center}
\begin{picture}(65,44)

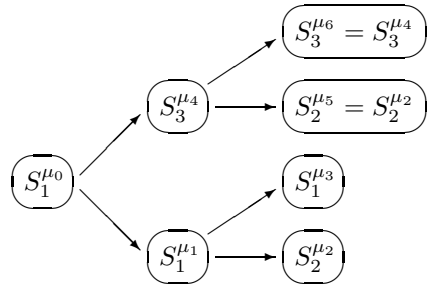
\put(00,00){\makebox(0,0)[bl]{Fig. 21. Resultant tree-like
 trajectory}}

\put(4,19){\oval(8,7)}
\put(01.5,17){\makebox(0,8)[bl]{\(S^{\mu_{0}}_{1}\)}}

\put(09,18){\vector(1,-1){08}}

\put(09,20){\vector(1,1){08}}


\put(22,09){\oval(8,7)}
\put(19.5,07){\makebox(0,8)[bl]{\(S^{\mu_{1}}_{1}\)}}

\put(27,09){\vector(1,0){08}}

\put(26,12){\vector(3,2){09}}


\put(40,09){\oval(8,7)}
\put(37.5,07){\makebox(0,8)[bl]{\(S^{\mu_{2}}_{2}\)}}


\put(40,19){\oval(8,7)}
\put(37.5,17){\makebox(0,8)[bl]{\(S^{\mu_{3}}_{1}\)}}



\put(22,29){\oval(8,7)}
\put(19.5,27){\makebox(0,8)[bl]{\(S^{\mu_{4}}_{3}\)}}

\put(27,29){\vector(1,0){08}}

\put(26,32){\vector(3,2){09}}



\put(45.5,29){\oval(19,7)}
\put(37.5,27){\makebox(0,8)[bl]{\(S^{\mu_{5}}_{2} =S^{\mu_{2}}_{2}
\)}}


\put(45.5,39){\oval(19,7)}
\put(37.5,37){\makebox(0,8)[bl]{\(S^{\mu_{6}}_{3}=S^{\mu_{4}}_{3}\)}}

\end{picture}
\end{center}

\subsection{Simplified Example over Directed Graph}

 Here,
 a simplified example based on directed graph for top-level
 network is presented.
 This is a transformation of the example from previous section:~
 medical treatment.
 Table 13 and Table 14 contains descriptions
 of logical points and 'analysis/decision' points.
 The same morphological structures of the treatment plans
 are examined.
 The decision tree from Fig. 15 is transformed into a general graph
 (with feedbacks): Fig. 22.
 In  Fig. 23, an example of a preliminary solution graph is presented
 (e.g., for a certain patient).

\begin{center}
\begin{picture}(60,62)

\put(09,58){\makebox(0,0)[bl] {Table 13. Logical points}}

\put(00,00){\line(1,0){58}} \put(0,46.5){\line(1,0){58}}
\put(00,56){\line(1,0){58}}

\put(00,00){\line(0,1){56}} \put(13,00){\line(0,1){56}}
\put(58,00){\line(0,1){56}}


\put(01,51.8){\makebox(0,0)[bl] {Logical}}
\put(01,48){\makebox(0,0)[bl] {point}}

\put(26,52){\makebox(0,0)[bl] {Description}}


\put(01,42){\makebox(0,0)[bl]{\(\mu_{0}\)}}
\put(14,41.5){\makebox(0,0)[bl]{Basic (fool) treatment}}
\put(14,38){\makebox(0,0)[bl]{(Fig. 16)}}


\put(01,34){\makebox(0,0)[bl]{\(\mu_{1}\)}}
\put(14,34){\makebox(0,0)[bl]{Treatment by environment}}
\put(14,30){\makebox(0,0)[bl]{(Fig. 17)}}


\put(01,26){\makebox(0,0)[bl]{\(\mu_{2}\)}}

\put(14,26){\makebox(0,0)[bl]{Additional treatment}}
\put(14,22){\makebox(0,0)[bl]{by relaxation (Fig. 18)}}


\put(01,18){\makebox(0,0)[bl]{\(\mu_{3}\)}}
\put(14,18){\makebox(0,0)[bl]{Additional physical therapy}}
\put(14,14){\makebox(0,0)[bl]{(Fig. 19)}}


\put(01,10){\makebox(0,0)[bl]{\(\mu_{4}\)}}

\put(14,10){\makebox(0,0)[bl]{Additional physical therapy}}
\put(14,06){\makebox(0,0)[bl]{and drug treatment}}
\put(14,02){\makebox(0,0)[bl]{(Fig. 20)}}





\end{picture}
%
\begin{picture}(56,69)


\put(00.5,65){\makebox(0,0)[bl] {Table 14. 'Analysis/decision'
 points}}

\put(00,00){\line(1,0){56}} \put(00,49.8){\line(1,0){56}}
\put(00,63){\line(1,0){56}}

\put(00,00){\line(0,1){63}} \put(16.5,00){\line(0,1){63}}
\put(56,00){\line(0,1){63}}


\put(0.5,58.6){\makebox(0,0)[bl] {'Analysis/}}
\put(01,55.3){\makebox(0,0)[bl] {decision'}}
\put(01,51){\makebox(0,0)[bl] {point}}

\put(25,58.6){\makebox(0,0)[bl] {Description}}


\put(01,44){\makebox(0,0)[bl]{\(a_{0}\)}}

\put(17,44){\makebox(0,0)[bl]{(i) good results, go to
 \(\mu_{1}\)}}

\put(17,40){\makebox(0,0)[bl]{(ii) not sufficient results,}}
\put(22,36.5){\makebox(0,0)[bl]{go to \(\mu_{4}\)}}


\put(01,31){\makebox(0,0)[bl]{\(a_{1}\)}}

\put(17,31){\makebox(0,0)[bl]{(i) good results, go to
 \(\mu_{2}\)}}

\put(17,27){\makebox(0,0)[bl]{(ii) not sufficient results,}}
\put(22,23.5){\makebox(0,0)[bl]{go to \(\mu_{3}\)}}


\put(01,18){\makebox(0,0)[bl]{\(a_{4}\)}}

\put(17,18){\makebox(0,0)[bl]{(i) good results, go to
 \(\mu_{2}\)}}

\put(17,14){\makebox(0,0)[bl]{(ii) medium level of}}

\put(22,10){\makebox(0,0)[bl]{results, go to \(\mu_{1}\)}}

\put(17,06){\makebox(0,0)[bl]{(ii) not sufficient results,}}
\put(22,02.5){\makebox(0,0)[bl]{go to \(\mu_{0}\)}}

\end{picture}
\end{center}

\begin{center}
\begin{picture}(76,45)

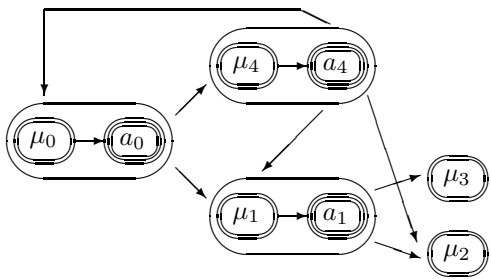
\put(04.2,00){\makebox(0,0)[bl]{Fig. 22. Example of general
 graph}}


\put(11,24){\oval(22,10)}

\put(05,24){\oval(8,6)} \put(05,24){\oval(7,5)}
\put(03,23){\makebox(0,8)[bl]{\(\mu_{0}\)}}

\put(09,24){\vector(1,0){4}}


\put(17,24){\oval(8,6)} \put(17,24){\oval(7,5)}
\put(17,24){\oval(6,4)}

\put(15,23){\makebox(0,8)[bl]{\(a_{0}\)}}

\put(22.5,27.5){\vector(1,1){4}} \put(22.5,20.5){\vector(1,-1){4}}


\put(38,14){\oval(22,10)}

\put(32,14){\oval(8,6)} \put(32,14){\oval(7,5)}
\put(30,13){\makebox(0,8)[bl]{\(\mu_{1}\)}}

\put(36,14){\vector(1,0){4}}


\put(44,14){\oval(8,6)} \put(44,14){\oval(7,5)}
\put(44,14){\oval(6,4)}

\put(42,13){\makebox(0,8)[bl]{\(a_{1}\)}}

\put(49,17.5){\vector(3,1){6}} \put(49,10.5){\vector(3,-1){6}}


\put(60,09){\oval(8,6)} \put(60,09){\oval(7,5)}
\put(58,08){\makebox(0,8)[bl]{\(\mu_{2}\)}}


\put(60,19){\oval(8,6)} \put(60,19){\oval(7,5)}
\put(58,18){\makebox(0,8)[bl]{\(\mu_{3}\)}}



\put(38,34){\oval(22,10)}

\put(32,34){\oval(8,6)} \put(32,34){\oval(7,5)}
\put(30,33){\makebox(0,8)[bl]{\(\mu_{4}\)}}

\put(36,34){\vector(1,0){4}}


\put(44,34){\oval(8,6)} \put(44,34){\oval(7,5)}
\put(44,34){\oval(6,4)}

\put(42,33){\makebox(0,8)[bl]{\(a_{4}\)}}

\put(43,39.5){\line(-2,1){4}} \put(39,41.5){\line(-1,0){34}}

\put(05,41.5){\vector(0,-1){11.4}}

\put(42,28){\vector(-1,-1){8}}

\put(48,30){\vector(1,-3){6.5}}








\end{picture}
%
\begin{picture}(64,45)

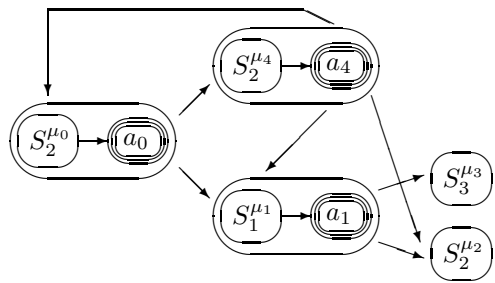
\put(03.2,00){\makebox(0,0)[bl]{Fig. 23. Example of solution
 graph}}


\put(11,24){\oval(22,10)}

\put(05,24){\oval(8,7)}


\put(02.5,22){\makebox(0,8)[bl]{\(S^{\mu_{0}}_{2}\)}}

\put(09,24){\vector(1,0){4}}


\put(17,24){\oval(8,6)} \put(17,24){\oval(7,5)}
\put(17,24){\oval(6,4)}

\put(15,23){\makebox(0,8)[bl]{\(a_{0}\)}}

\put(22.5,27.5){\vector(1,1){4}} \put(22.5,20.5){\vector(1,-1){4}}


\put(38,14){\oval(22,10)}

\put(32,14){\oval(8,7)}


\put(29.5,12){\makebox(0,8)[bl]{\(S^{\mu_{1}}_{1}\)}}

\put(36,14){\vector(1,0){4}}


\put(44,14){\oval(8,6)} \put(44,14){\oval(7,5)}
\put(44,14){\oval(6,4)}

\put(42,13){\makebox(0,8)[bl]{\(a_{1}\)}}

\put(49,17.5){\vector(3,1){6}} \put(49,10.5){\vector(3,-1){6}}


\put(60,09){\oval(8,7)}


\put(57.5,07){\makebox(0,8)[bl]{\(S^{\mu_{2}}_{2}\)}}


\put(60,19){\oval(8,7)}


\put(57.5,17){\makebox(0,8)[bl]{\(S^{\mu_{3}}_{3}\)}}



\put(38,34){\oval(22,10)}

\put(32,34){\oval(8,7)}


\put(29.5,32){\makebox(0,8)[bl]{\(S^{\mu_{4}}_{2}\)}}

\put(36,34){\vector(1,0){4}}


\put(44,34){\oval(8,6)} \put(44,34){\oval(7,5)}
\put(44,34){\oval(6,4)}

\put(42,33){\makebox(0,8)[bl]{\(a_{4}\)}}

\put(43,39.5){\line(-2,1){4}} \put(39,41.5){\line(-1,0){34}}

\put(05,41.5){\vector(0,-1){11.4}}

\put(42,28){\vector(-1,-1){8}}

\put(48,30){\vector(1,-3){6.5}}








\end{picture}
\end{center}

\subsection{On Multiple Domain Problems}

 The multistage strategies can be considered for various domains:
  system design, system testing, medical
 treatment, medical diagnosis.
 Fig. 24 illustrates a multistage trajectory for multistage  diagnosis
 or system testing:~

 (i)  points \(\{ \theta_{0},\theta_{1},...,\theta_{k} \}\)
 correspond to diagnosis/testing,

 (ii) points \(\{ a_{0},a_{1},...,a_{k}\}\)
 correspond to analysis/decision, and

 (iii)  morphological structures for
 test points are:
 \(\{\Lambda^{\theta_{0}},\Lambda^{\theta_{1}},...,\Lambda^{\theta_{k}}\}\).

 On the other hand,
 it may be reasonable to examine
 multistage trajectories for two domains (Fig. 25):
 (a) system testing, (b) system design.
%
%
 Here, the following notations are used:

%
 (i) points \(\{ \theta_{1},\theta_{2},...,\theta_{k}\}\)
 correspond to system testing/diagnosis,

 (ii) points \(\{ a_{0},a_{1},...,a_{k}\}\)
 correspond to analysis/decision,

 (iii) points \(\{ \xi_{1},\xi_{2},...,\xi_{q}\}\)
 correspond to system design/redesign,

 (iv) morphological structures for
 test points are:
 \(\{\Lambda^{\theta_{0}},\Lambda^{\theta_{1}},...,\Lambda^{\theta_{k}}\}\),
 and

 (v)  morphological structures for
 design/redesign points are:
 \(\{\Lambda^{\xi_{0}},\Lambda^{\xi_{1}},...,\Lambda^{\xi_{q}}\}\).

\begin{center}
\begin{picture}(95,35)

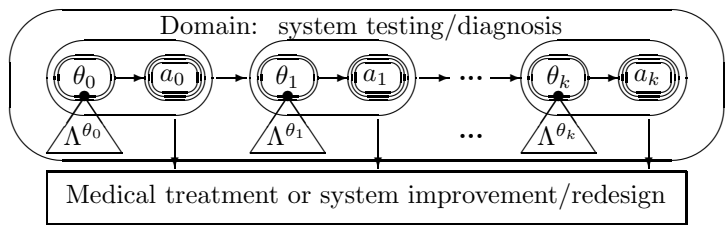
\put(09,00){\makebox(0,0)[bl]{Fig. 24. Example of multistage
 testing/treatment}}

\put(20,30){\makebox(0,0)[bl]{Domain: ~system testing/diagnosis}}
\put(47.5,24){\oval(95,20)}


\put(16,25){\oval(22,10)}

\put(10,25){\oval(8,6)} \put(10,25){\oval(7,5)}
\put(08.4,23.7){\makebox(0,8)[bl]{\(\theta_{0}\)}}

\put(14,25){\vector(1,0){4}}


\put(10,22.5){\circle*{1.5}}

\put(05,15){\line(2,3){5}} \put(15,15){\line(-2,3){5}}
\put(05,15){\line(1,0){10}}

\put(07.2,16){\makebox(0,8)[bl]{\(\Lambda^{\theta_{0}}\)}}


\put(22,25){\oval(8,6)} \put(22,25){\oval(7,5)}
\put(22,25){\oval(6,4)}

\put(20,24){\makebox(0,8)[bl]{\(a_{0}\)}}

\put(27.5,25){\vector(1,0){4}}

\put(22,19.5){\vector(0,-1){6.5}}


\put(43,25){\oval(22,10)}

\put(37,25){\oval(8,6)} \put(37,25){\oval(7,5)}
\put(35.4,23.7){\makebox(0,8)[bl]{\(\theta_{1}\)}}

\put(41,25){\vector(1,0){4}}


\put(37,22.5){\circle*{1.5}}

\put(32,15){\line(2,3){5}} \put(42,15){\line(-2,3){5}}
\put(32,15){\line(1,0){10}}

\put(34.2,16){\makebox(0,8)[bl]{\(\Lambda^{\theta_{1}}\)}}


\put(49,25){\oval(8,6)} \put(49,25){\oval(7,5)}
\put(49,25){\oval(6,4)}

\put(47,24){\makebox(0,8)[bl]{\(a_{1}\)}}

\put(54.5,25){\vector(1,0){4}}

\put(49,19.5){\vector(0,-1){6.5}}


\put(59.7,24.7){\makebox(0,8)[bl]{{\bf ...}}}

\put(64,25){\vector(1,0){4}}

\put(59.7,17){\makebox(0,8)[bl]{{\bf ...}}}


\put(79,25){\oval(22,10)}

\put(73,25){\oval(8,6)} \put(73,25){\oval(7,5)}
\put(71.4,23.7){\makebox(0,8)[bl]{\(\theta_{k}\)}}

\put(77,25){\vector(1,0){4}}


\put(73,22.5){\circle*{1.5}}

\put(68,15){\line(2,3){5}} \put(78,15){\line(-2,3){5}}
\put(68,15){\line(1,0){10}}

\put(70.2,16){\makebox(0,8)[bl]{\(\Lambda^{\theta_{k}}\)}}


\put(85,25){\oval(8,6)} \put(85,25){\oval(7,5)}
\put(85,25){\oval(6,4)}

\put(83,24){\makebox(0,8)[bl]{\(a_{k}\)}}

\put(85,19.5){\vector(0,-1){6.5}}


\put(07.5,07.5){\makebox(0,8)[bl]{Medical treatment or
 system improvement/redesign}}

\put(05,05.5){\line(1,0){85}} \put(05,12.5){\line(1,0){85}}

\put(05,05.5){\line(0,1){7}} \put(90,05.5){\line(0,1){7}}

\end{picture}
\end{center}

\begin{center}
\begin{picture}(95,48)

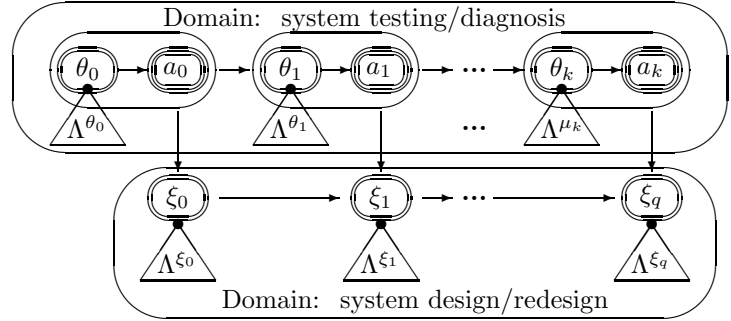
\put(00,00){\makebox(0,0)[bl]{Fig. 25. Two-domain multistage
 trajectory  (testing\&design)}}

\put(20,43){\makebox(0,0)[bl]{Domain: ~system testing/diagnosis}}
\put(47.5,37){\oval(95,20)}


\put(16,38){\oval(22,10)}

\put(10,38){\oval(8,6)} \put(10,38){\oval(7,5)}
\put(08.4,36.7){\makebox(0,8)[bl]{\(\theta_{0}\)}}

\put(14,38){\vector(1,0){4}}


\put(10,35.5){\circle*{1.5}}

\put(05,28){\line(2,3){5}} \put(15,28){\line(-2,3){5}}
\put(05,28){\line(1,0){10}}

\put(07.2,29){\makebox(0,8)[bl]{\(\Lambda^{\theta_{0}}\)}}


\put(22,38){\oval(8,6)} \put(22,38){\oval(7,5)}
\put(22,38){\oval(6,4)}

\put(20,37){\makebox(0,8)[bl]{\(a_{0}\)}}

\put(27.5,38){\vector(1,0){4}}

\put(22,32.5){\vector(0,-1){8}}


\put(43,38){\oval(22,10)}

\put(37,38){\oval(8,6)} \put(37,38){\oval(7,5)}
\put(35.4,36.7){\makebox(0,8)[bl]{\(\theta_{1}\)}}

\put(41,38){\vector(1,0){4}}


\put(37,35.5){\circle*{1.5}}

\put(32,28){\line(2,3){5}} \put(42,28){\line(-2,3){5}}
\put(32,28){\line(1,0){10}}

\put(34.2,29){\makebox(0,8)[bl]{\(\Lambda^{\theta_{1}}\)}}


\put(49,38){\oval(8,6)} \put(49,38){\oval(7,5)}
\put(49,38){\oval(6,4)}

\put(47,37){\makebox(0,8)[bl]{\(a_{1}\)}}

\put(54.5,38){\vector(1,0){4}}

\put(49,32.5){\vector(0,-1){8}}


\put(59.7,37.7){\makebox(0,8)[bl]{{\bf ...}}}

\put(64,38){\vector(1,0){4}}

\put(59.7,30){\makebox(0,8)[bl]{{\bf ...}}}


\put(79,38){\oval(22,10)}

\put(73,38){\oval(8,6)} \put(73,38){\oval(7,5)}
\put(71.4,36.7){\makebox(0,8)[bl]{\(\theta_{k}\)}}

\put(77,38){\vector(1,0){4}}


\put(73,35.5){\circle*{1.5}}

\put(68,28){\line(2,3){5}} \put(78,28){\line(-2,3){5}}
\put(68,28){\line(1,0){10}}

\put(70.2,29){\makebox(0,8)[bl]{\(\Lambda^{\mu_{k}}\)}}


\put(85,38){\oval(8,6)} \put(85,38){\oval(7,5)}
\put(85,38){\oval(6,4)}

\put(83,37){\makebox(0,8)[bl]{\(a_{k}\)}}

\put(85,32.5){\vector(0,-1){8}}


\put(22,21){\oval(8,6)} \put(22,21){\oval(7,5)}
\put(20.4,19.5){\makebox(0,8)[bl]{\(\xi_{0}\)}}


\put(22,17.5){\circle*{1.5}}

\put(17,10){\line(2,3){5}} \put(27,10){\line(-2,3){5}}
\put(17,10){\line(1,0){10}}

\put(19.2,11){\makebox(0,8)[bl]{\(\Lambda^{\xi_{0}}\)}}


\put(27.5,21){\vector(1,0){16}}


\put(49,21){\oval(8,6)} \put(49,21){\oval(7,5)}
\put(47.4,19.5){\makebox(0,8)[bl]{\(\xi_{1}\)}}


\put(49,17.5){\circle*{1.5}}

\put(44,10){\line(2,3){5}} \put(54,10){\line(-2,3){5}}
\put(44,10){\line(1,0){10}}

\put(46.2,11){\makebox(0,8)[bl]{\(\Lambda^{\xi_{1}}\)}}


\put(54.5,21){\vector(1,0){4}}


\put(59.7,20.7){\makebox(0,8)[bl]{{\bf ...}}}

\put(64,21){\vector(1,0){16}}


\put(85,21){\oval(8,6)} \put(85,21){\oval(7,5)}
\put(83.4,19.5){\makebox(0,8)[bl]{\(\xi_{q}\)}}


\put(85,17.5){\circle*{1.5}}

\put(80,10){\line(2,3){5}} \put(90,10){\line(-2,3){5}}
\put(80,10){\line(1,0){10}}

\put(82.2,11){\makebox(0,8)[bl]{\(\Lambda^{\xi_{q}}\)}}


\put(27.8,05.5){\makebox(0,0)[bl]{Domain: ~system
 design/redesign}}

\put(53.5,15){\oval(80,20)}

\end{picture}
\end{center}
%

%
 Thus, the combined multistage strategy involves two parts:
 (a) multistage strategy for system testing and
 (b) multistage strategy for system improvement/redesign.
 Note,
 two-domain problem can correspond to the following domains:
 system utilization and system maintenance.
 Evidently, parts for different domains  can have more complicated
 forms (i.e., over digraph).


\section{Conclusion}

 The paper describes multistage design for a composite (modular)
 system (i.e., design of system trajectories).
 In the complicated situations,
 the multistage design is extended
 for graph-based design structures.
%
 Composition of the graph-based design structures
 lead to prospective multi-domain problems, for example:
%
 (a) system testing/diagnosis and system improvement/redesign,
 (b) medical diagnosis and medical treatment,
 (c) system utilization and system maintenance.
 It is necessary to note,
 the described approach is close to finite-state machines
 or state transition diagrams
 (e.g., \cite{dru06,harel87,harel98,hopcroft79,jac83,was85}).
 On the other hand, it is interesting to consider a similarity
 of
 the described approach and dynamic decision making methods
 (e.g., \cite{da00}).
%

%
 In the future, it may be  reasonable to consider
 the following research directions:

 (1) examination of various real-world applications
 (e.g., medical diagnosis and medical treatment,
 communication networks);

 (2) study and usage of multistage system trajectories
 for multiple domains problems
 (e.g., system testing, system maintenance, system utilization, system improvement/redesign)
 including inter-domains interconnection;
%
%

 (3) special study of formal models
  for 'analysis/decision' points (states);

 (4) taking into account uncertainty
 and extension of the described approach;

 (5) examination of a dynamic described approach
 while taking into account changes of external requirements;

 (6) design of a special computer environment;
 and

 (7) usage of the described system approaches in
 education (computer science, engineering, applied mathematics,
 management).

\end{document}